%% file: main.tex
\documentclass[runningheads]{llncs}

\usepackage{eccv}

\usepackage{eccvabbrv}

\usepackage{graphicx}
\usepackage{booktabs}

\usepackage[accsupp]{axessibility}  % Improves PDF readability for those with disabilities.

\input{preamble}

\usepackage[pagebackref,breaklinks,colorlinks,citecolor=eccvblue]{hyperref}
\usepackage{orcidlink}

\usepackage{wrapfig}

\begin{document}

\title{\ours: Reasoning Over Object Tracks for Zero-Shot Referring Video Object Segmentation}

\titlerunning{\ours}

% \author{Woojeong Jin\inst{1} \and
% Jaeho Lee\inst{1} \and
% Heeseong Shin\inst{1} \and \\
% Seungho Jang\inst{2} \and
% Junhwan Heo\inst{3} \and
% Seungryong Kim\inst{1}$^{,\dagger}$}
\author{Woojeong Jin$^*$ \and
Jaeho Lee$^*$ \and
Heeseong Shin \and \\
Seungho Jang \and
Junhwan Heo \and
Seungryong Kim$^{\dagger}$}

\authorrunning{Jin et al.}

\institute{KAIST AI \\[1em]
Project page: \url{https://cvlab-kaist.github.io/AgentRVOS} \\
}

\maketitle

\input{fig/teaser}

\input{sec/0_abstract}
\input{sec/1_introduction}
\input{sec/2_related_work}
\input{sec/3_method}
\input{sec/4_experiments}
\input{sec/5_conclusion}

\clearpage

\input{sec/suppl}

% \section*{Acknowledgements}
% Please insert your acknowledgments here.

\clearpage

\bibliographystyle{splncs04}
\bibliography{main}
\end{document}

%% file: preamble.tex
\usepackage{amsfonts}
\usepackage{amsmath}
\usepackage{multirow}
\usepackage{amssymb}
\usepackage{pifont}
\usepackage{lipsum}
\usepackage{xspace}
\usepackage{mwe}
\usepackage{colortbl}
\usepackage{subcaption}

\usepackage{algorithm}
\usepackage{algpseudocode}
\usepackage{placeins}

\usepackage{array}
\newcolumntype{L}[1]{>{\raggedright\let\newline\\\arraybackslash\hspace{0pt}}m{#1}}
\newcolumntype{Z}[1]{>{\centering\let\newline\\\arraybackslash\hspace{0pt}}m{#1}}
\newcolumntype{R}[1]{>{\raggedleft\let\newline\\\arraybackslash\hspace{0pt}}m{#1}}

\newcommand{\ours}{AgentRVOS\xspace}

\definecolor{qrand}{HTML}{2b2e4a}
\definecolor{qtext}{HTML}{e84545}
\definecolor{blue}{HTML}{e8f0f8}

%% file: fig/teaser.tex
\begin{figure*}[!h]
    \centering
    \vspace{-20pt}
    \includegraphics[width=\textwidth]{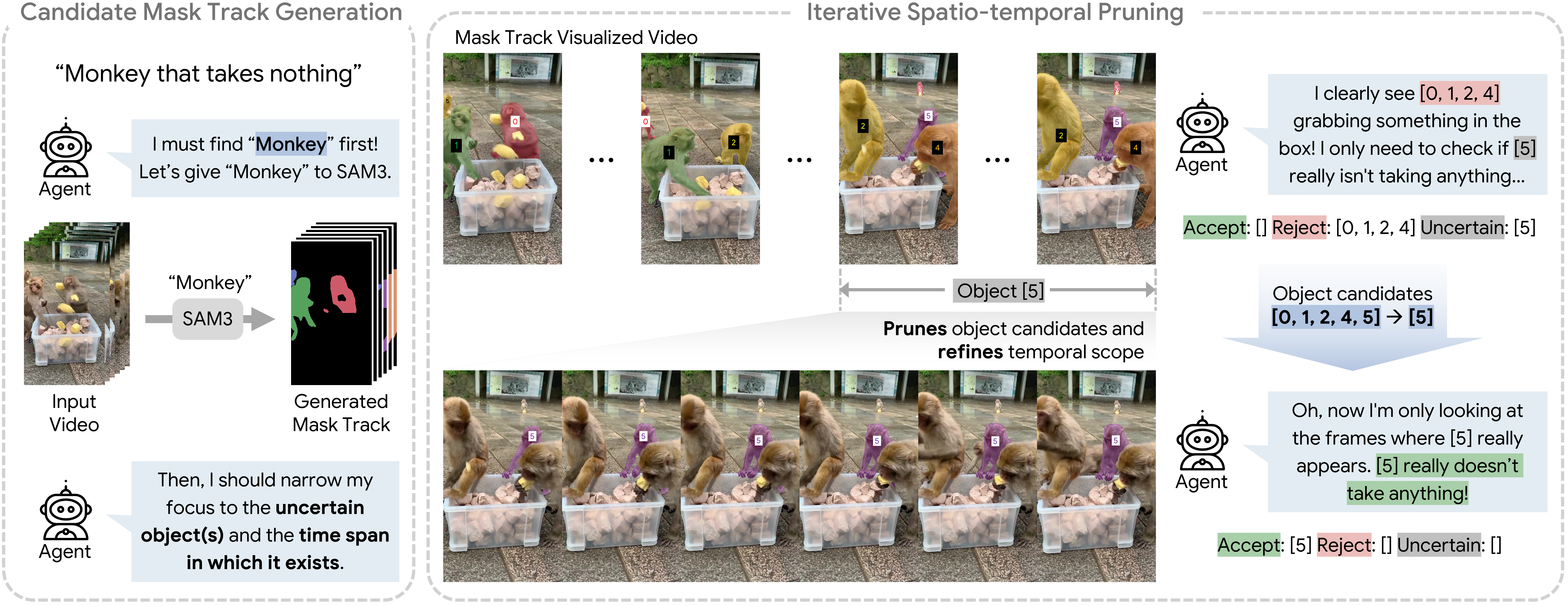}
    \caption{\textbf{Teaser.}
    \ours is a training-free agentic pipeline built on the complementary strengths of SAM3~\cite{carion2025sam3} and an MLLM~\cite{bai2025qwen3, openai2025gpt5}. The MLLM first uses SAM3 to generate candidate mask tracks, then iteratively prunes them through query-grounded reasoning over object-level evidence.
    }
    \label{fig:teaser}
    \vspace{-25pt}
\end{figure*}

%% file: sec/0_abstract.tex
\begin{abstract}
    Referring Video Object Segmentation (RVOS) aims to segment a target object throughout a video given a natural language query. Training-free methods for this task follow a common pipeline: an MLLM selects keyframes, grounds the referred object within those frames, and a video segmentation model propagates the results. While intuitive, this design asks the MLLM to make temporal decisions before any object-level evidence is available, limiting both reasoning quality and spatio-temporal coverage.
    To overcome this, we propose \textbf{\ours}, a training-free agentic pipeline built on the complementary strengths of SAM3 and an MLLM.
    Given a concept derived from the query, SAM3 provides reliable perception over the full spatio-temporal extent through generated mask tracks.
    The MLLM then identifies the target through query-grounded reasoning over this object-level evidence, iteratively pruning guided by SAM3's temporal existence information. Extensive experiments show that \ours achieves state-of-the-art performance among training-free methods across multiple benchmarks, with consistent results across diverse MLLM backbones.
    \keywords{Referring Video Object Segmentation \and Agentic AI}
\end{abstract}

%% file: sec/1_introduction.tex
\section{Introduction}
\label{sec:intro}

% P1: Task Definition + Core challenges
Referring Video Object Segmentation (RVOS) requires generating the segmentation mask tracks of the target object throughout a video based on a given natural language query. Unlike image-level referring segmentation~\cite{lai2024lisa, kao2026cotsegrethinkingsegmentationchainofthought, liu2025segzeroreasoningchainguidedsegmentation}, RVOS involves queries that go beyond static appearance descriptions and possesses video-specific challenges such as temporal ordering, complex motions, and inter-object relations~\cite{ding2023mevis, yan2024visa, jin2025interrvos}. These challenges give rise to two intertwined requirements. First, the model must reason about complex temporal and relational queries that distinguish a specific object among many based on actions, state changes, or inter-object relations. Second, it must ensure dense spatio-temporal coverage, as target objects may be small, non-salient, or appear only briefly within a long sequence.

% P3: Training-free MLLM Approaches
With the recent advances in multimodal large language models (MLLMs)~\cite{liu2023visual, li2024llava, team2023gemini, hurst2024gpt}, the strong reasoning capabilities of these billion-scale models have shown significant promise for the RVOS task, particularly in comprehending the complex queries given in the task.
Existing approaches adopt MLLMs either through task-specific fine-tuning~\cite{yuan2025sa2va, bai2024videolisa, yan2024visa, lin2025glus}, or more recently, in a training-free manner that directly leverages their native multi-modal reasoning capabilities over images and videos.

For the training-free methods~\cite{huang2025alrefsam2, kao2025cotrvs, jiang2026referagent}, a common approach is to first identify a set of video frames that are relevant to a given query, and then perform object grounding on this set.
Then, a video segmentation model, such as segment anything 2 (SAM2)~\cite{ravi2024sam2}, propagates the initial masks across the remaining frames to produce the full segmentation across the entire video. Consequently, these pipelines \textbf{heavily rely on the MLLM for both temporal frame selection and spatial grounding}. However, MLLMs often operate on sparsely sampled frames due to input token limits, resulting in limited temporal coverage. This makes it difficult to detect objects that appear only briefly or intermittently in long videos, suggesting that offloading precise spatio-temporal perception (\textit{i.e.}, temporal object detection) from the MLLM would allow the model to focus entirely on its primary strength such as complex reasoning.

\input{fig/motivation}

% P5: Our Approach: SAM3 + MLLM Complementary
In this paper, we present \textbf{\ours}, a \textbf{training-free agentic} pipeline for RVOS that leverages the reasoning capabilities of MLLMs to their fullest extent by incorporating SAM3~\cite{carion2025sam3} as a complementary perceptual tool.
Given a textual prompt, SAM3 can process all frames of a video and produce high quality mask tracks for every matching object instance, without the need for additional inputs such as points or bounding boxes. This allows us to reliably detect small, occluded objects, while also recognizing briefly appearing objects that MLLMs often overlook, as SAM3 can examine the entire video rather than sparse samples. Consequently, we can identify exactly which frames a given object appears in with frame-level precision, enabling us to leverage the MLLM to reason within this spatio-temporally constrained segment.

However, SAM3 alone is not sufficient for RVOS, as it is designed to accept concepts--often given as short noun phrases (\eg, ``\textit{person}'', ``\textit{red car}'')--as inputs, and tends to struggle with complex queries including temporal or relational comprehension.
For instance, given ``\textit{the person who stands up after sitting}'', SAM3 can easily locate each person in the video, but cannot determine which one exactly is exhibiting the described behavior.
This is where the reasoning capability of MLLMs becomes essential. Given the candidate tracks produced by SAM3, the MLLM determines which one corresponds to the target by performing fine-grained temporal and relational reasoning over the full sentence query.
In this way, SAM3 and MLLM complement each other: \textbf{SAM3} provides reliable \textbf{perception over the full spatio-temporal extent} of the video, while the \textbf{MLLM} contributes \textbf{comprehensive query-grounded reasoning} over the resultant object-level evidence, as illustrated in Fig.~\ref{fig:teaser}.

% P6: Iterative Pipeline
Translating this complementary structure into practice requires addressing an additional challenge: SAM3's concept-level candidate generation is intentionally exhaustive, producing a large and diverse candidate pool to ensure high recall.
Presenting all candidates to the MLLM at once, however, is impractical, as overlapping mask visualizations may degrade reasoning quality and candidates may span different temporal intervals.

We therefore design \ours as an iterative agentic pipeline that progressively narrows both the candidate set and the temporal scope. At each iteration, MLLM progressively accepts or rejects candidates based on available evidence, while SAM3's temporal existence information guides focused re-examination of the remaining uncertain cases.
Through this progressive narrowing--fewer candidates, tighter temporal windows, and simpler reasoning at each iteration--the pipeline converges on the target object without requiring an explicit frame selection module or exhaustive processing of the entire video.

% P7: Contributions
Our contributions are summarized as follows:
\begin{itemize}
    \item We propose \textbf{\ours}, a training-free agentic pipeline for RVOS that combines SAM3's language-grounded perception with the MLLM's reasoning capabilities. By delegating object detection and temporal localization to SAM3, our pipeline allows the MLLM to reason over structured, object-level evidence.
    \item We introduce an iterative spatio-temporal pruning strategy in which the MLLM progressively eliminates candidates while SAM3's temporal existence information narrows the relevant temporal scope, decomposing the complex selection problem into progressively simpler reasoning steps.
    \item Extensive experiments demonstrate that \ours achieves state-of-the-art performance across multiple benchmarks. Our pipeline consistently shows strong results with various open-source and closed-source MLLMs, demonstrating its generalizability.
\end{itemize}

%% file: fig/motivation.tex
\begin{figure*}[t]
    \centering
    \includegraphics[width=\textwidth]{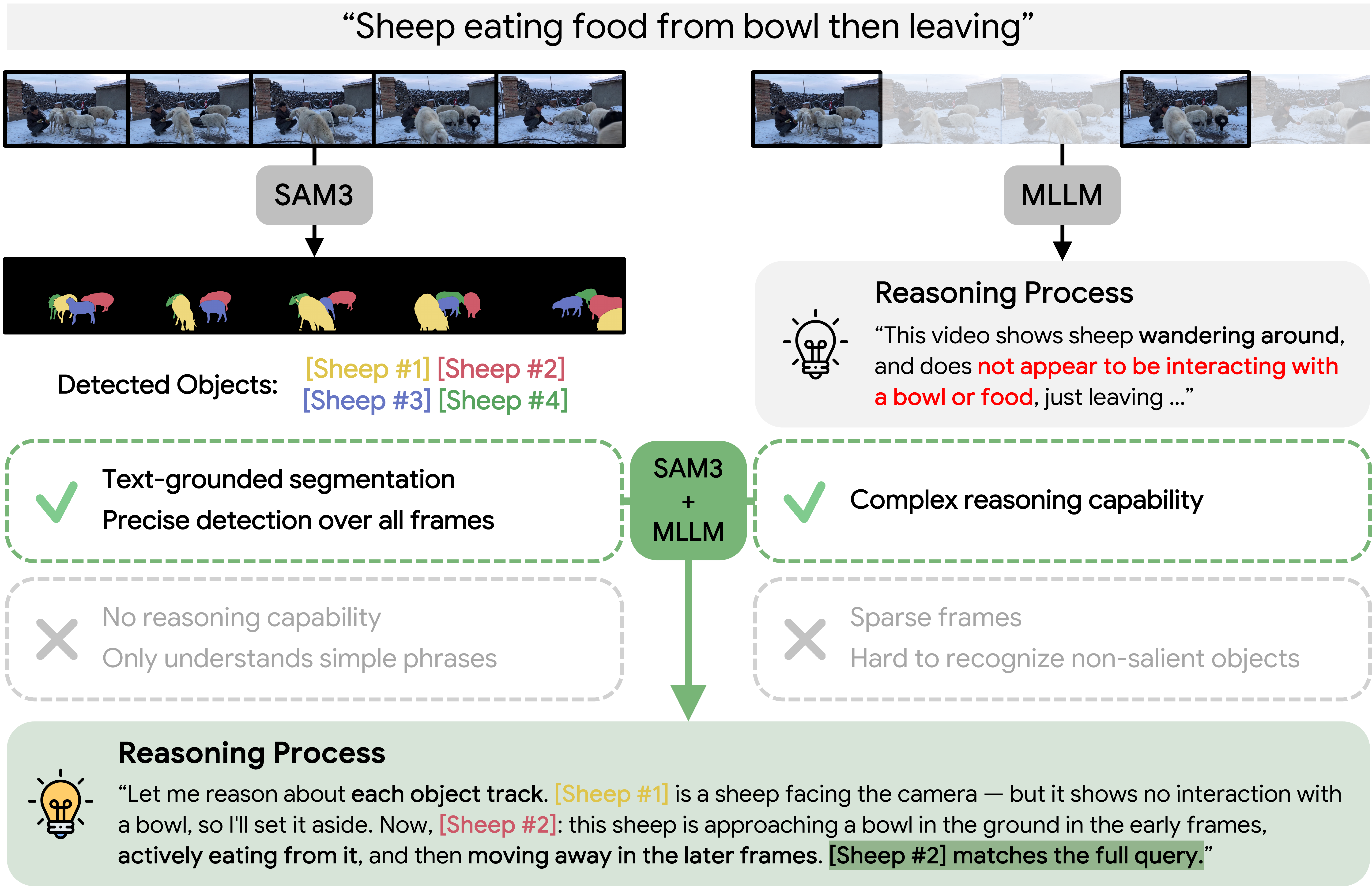}
    \caption{\textbf{Complementary concept of SAM3 and MLLM.}
    SAM3~\cite{carion2025sam3} can precisely identify objects without missing a single frame, but struggles with complex queries. MLLMs~\cite{bai2025qwen3, openai2025gpt5, li2024llava}, on the other hand, offer strong reasoning capabilities, but operate on sparse frames and struggle with non-salient objects. \ours combines the advantages of both SAM3 and MLLM, by interleaving the two models in a complementary manner.
    }
    \vspace{-15pt}
    \label{fig:sam3}
\end{figure*}

%% file: sec/2_related_work.tex
\section{Related Work}
\label{sec:related_work}

\vspace{-5pt}
\subsubsection{Referring Video Object Segmentation. }
\label{sec:related_rvos}
RVOS aims to segment target objects in a video based on natural language expressions. Earlier RVOS datasets~\cite{gavrilyuk2018a2dsentences, seo2020urvos, khoreva2018refvos} mainly focused on appearance-based expressions, where objects could be identified through static visual attributes. Consequently, pioneering works~\cite{seo2020urvos, wu2022referformer, he2024decouplingstatichierarchicalmotion, carion2020end, miao2024htr} demonstrated that query-based architectures--built upon DETR~\cite{carion2020end}--can effectively link appearance-based textual descriptions with visual representations of objects. However, more recent benchmarks introduce more challenging queries that go beyond appearance-based descriptions. For example, MeViS~\cite{ding2023mevis} introduces motion-centric expressions that require temporal reasoning and, ReVOS~\cite{yan2024visa} requires strong reasoning capabilities and world knowledge. These emerging datasets highlight the need for reasoning capabilities beyond simple appearance matching.

\input{fig/method}

\subsubsection{MLLM-based Reasoning Video Object Segmentation.}
To enhance reasoning ability for more and more challenging RVOS datasets, recent methods incorporate multimodal large language models (MLLMs)~\cite{li2024llava, chen2025expandingperformanceboundariesopensource, qwen2025qwen25technicalreport, bai2025qwen3} with foundation segmentation models~\cite{kirillov2023sam ,ravi2024sam2} through large-scale supervised fine-tuning~\cite{yan2024visa, bai2024videolisa, yuan2025sa2va} or reinforcement learning based optimization~\cite{xu2025videosegr1, li2025revsegincentivizingreasoningchain}. These approaches demonstrate strong performance and improved generalization to out-of-distribution samples.

However, such training-based paradigms typically require substantial annotated data and computational resources. To address these issues, training-free agentic approaches that leverage the zero-shot reasoning capability of MLLMs have recently emerged. 
For example, CoT-RVS~\cite{kao2025cotrvs} exploits the zero-shot Chain-of-Thought capability of MLLMs to select key frames, applies an image segmentation model to obtain masks for the target object on the selected frames, and propagates them across the video using a video processor. Similarly, Refer-Agent~\cite{jiang2026referagent} leverages CLIP~\cite{radford2021learning} and MLLMs for frame selection and utilizes MLLM-based grounding to prompt segmentation models.
Although these methods achieve competitive results without additional training, they rely heavily on the reasoning and grounding capabilities of MLLMs. MLLMs possess strong reasoning capabilities, but they typically process only a limited subset of frames due to token limitations. As a result, the model may struggle to perform reasoning centered on the referred object, particularly when the target object appears only sparsely over time. Moreover, the zero-shot grounding capability of MLLMs remains limited when handling small, blurred, or visually ambiguous objects, which can propagate errors across the video.

To address this limitation, we introduce \ours, a training-free agentic pipeline which incorporates SAM3 to provide dense spatio temporal object-level evidence across the entire video. This allows the MLLM to focus on query-grounded reasoning over reliable object-level cues.

%% file: fig/method.tex
\begin{figure*}[!t]
    \centering
    \includegraphics[width=\textwidth]{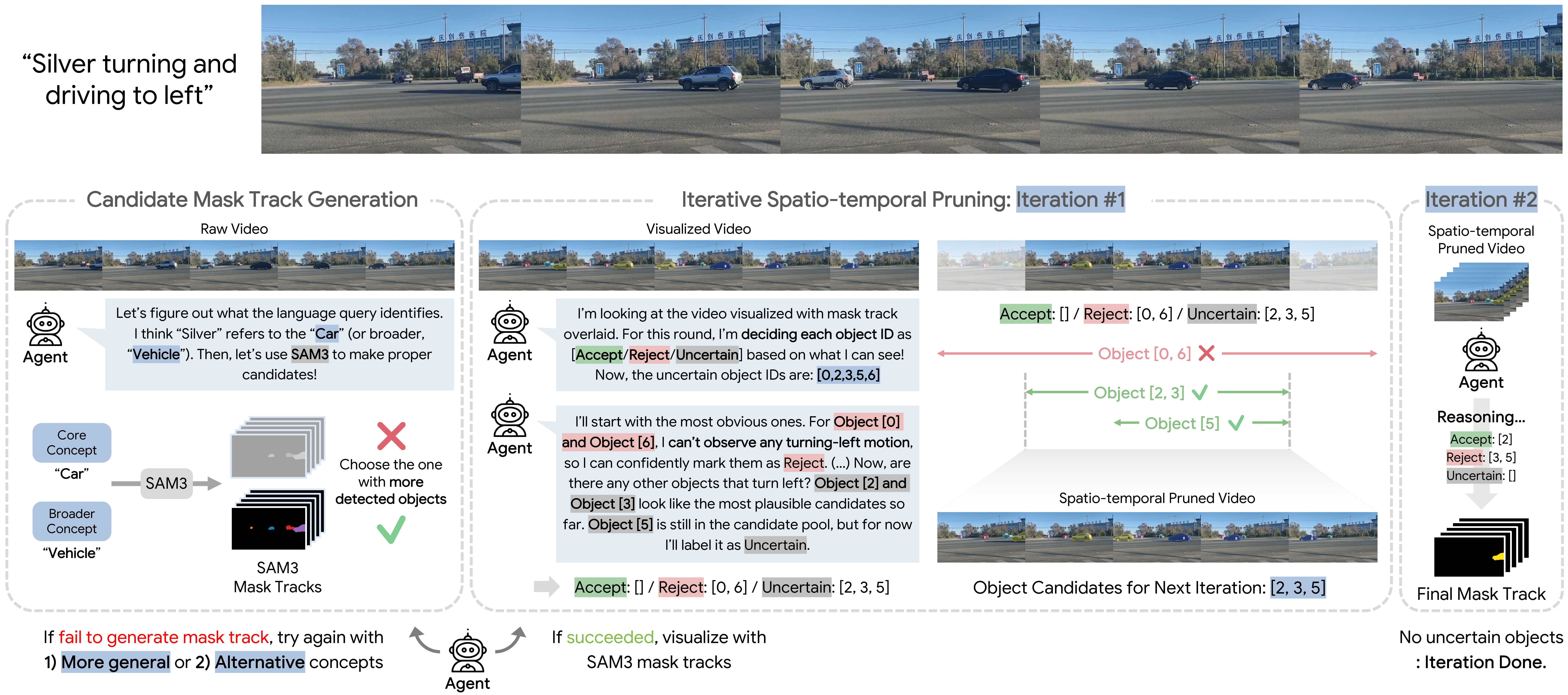}
    \caption{\textbf{Overall pipeline.}
    Given a video and a natural language query, our pipeline operates in two phases. In Candidate Mask Track Generation (Sec.~\ref{sec:method_candidate}), the MLLM first analyzes the query to extract concepts, which SAM3 uses to produce temporally consistent candidate mask tracks; this process iterates to ensure sufficient coverage. In Iterative Spatio-temporal Pruning (Sec.~\ref{sec:method_pruning}), the MLLM reasons over the candidate pool, classifying each candidate as \texttt{Accepted}, \texttt{Rejected}, or \texttt{Uncertain}, while progressively narrowing the spatio-temporal scope until convergence.
    }
    \vspace{-15pt}
    \label{fig:method}
\end{figure*}

\vspace{-5pt}

%% file: sec/3_method.tex
\section{Method}
\label{sec:method}

% 3.1 Problem Setting
\subsection{Problem Formulation and Overview}
Given a video $\mathcal{V} = \{I_t\}_{t=1}^{T}$ consisting of $T$ frames and a natural language query $Q$, Referring Video Object Segmentation (RVOS) aims to predict a sequence of $T$ binary masks $\mathcal{M} \in \{0, 1\}^{T \times H \times W}$ corresponding to the specified object, where $H$ and $W$ denote the spatial dimensions. To effectively tackle this, our approach leverages a synergistic integration of SAM3~\cite{carion2025sam3} and an MLLM~\cite{bai2025qwen3, openai2025gpt5}. The MLLM enhances SAM3's capabilities by translating the complex query $Q$ into a more interpretable prompt, while SAM3 reciprocates by extracting precise spatio-temporal mask tracks. These generated tracks serve as focused visual priors, enabling the MLLM to perform targeted reasoning on specific object candidates rather than exhaustively processing the entire video at once. 

Realizing this complementary pipeline, however, requires addressing two core challenges. First, the candidate mask tracks produced by SAM3 must actually contain the referred object; incomplete coverage at this stage cannot be recovered by later reasoning. Second, because SAM3 operates with limited semantic understanding of $Q$, necessitating a subsequent reasoning phase that can reliably distinguish the exact target.
Our method therefore proceeds in two phases: a candidate generation phase that prioritizes recall to ensure coverage, followed by a spatio-temporal pruning phase in which the MLLM leverages the object-level evidence provided by these tracks to identify and retain only the referred object.

In the first phase, \textbf{Candidate Mask Track Generation} (Sec.~\ref{sec:method_candidate}), the pipeline constructs a comprehensive pool of object candidates from the video.
The MLLM first analyzes the query $Q$ to determine whether it can be resolved from language alone (\ie, \textbf{referring}) or requires visual context from the video (\ie, \textbf{reasoning}). Based on this, the MLLM extracts concept-level inputs, especially noun-phrase inputs, and SAM3 then generates temporally consistent candidate mask tracks for each concept. This process iterates to ensure sufficient candidate coverage, expanding to broader or alternative concepts when the initial set is insufficient.
In the second phase, \textbf{Iterative Spatio-temporal Pruning} (Sec.~\ref{sec:method_pruning}), the MLLM performs query-grounded reasoning over the full candidate pool to identify the target.
At each iteration, the MLLM classifies candidates as \texttt{[Accepted, Rejected, Uncertain]}.
The temporal scope is then narrowed to focus on the frames where uncertain candidates exist, and the pruning repeats until no uncertain candidates remain. A detailed description of \ours is provided in Appendix~\ref{sec:suppl_concept_extraction}--\ref{sec:suppl_algorithm}.

% 3.3 Candidate Mask Track Generation
\subsection{Candidate Mask Track Generation}
\label{sec:method_candidate}
While it would be straightforward to directly predict mask tracks with SAM3 with the given query $Q$, SAM3 is trained to segment \textbf{concepts} - mainly referring to short noun phrases (\eg, ``\textit{person}'', ``\textit{red car}''), and tends to struggle with complex queries in RVOS, which requires comprehensive temporal or relational understanding. This motivates us to break down the complex query into simpler and more understandable concepts for SAM3 to handle.

\subsubsection{Concept Extraction.}
To this end, we first use the MLLM to pre-process the query $Q$ and extract a set of concepts $C$ that SAM3 can handle more reliably.
For \textbf{referring} queries, the target itself is self-contained in the query (\eg, ``\textit{the cat sitting on the red couch}'') and therefore can be identified from language alone without accessing the video.
For \textbf{reasoning} queries, where the target is defined through temporal or contextual cues that require visual understanding (\eg, ``\textit{the one that moves fastest}''), the object itself cannot be inferable solely from the language query. Thus, the MLLM examines sampled video frames alongside the query to infer the relevant object categories.

For extraction, we define two levels of granularity to ensure robustness: \textbf{core} concepts and \textbf{broader} concepts. \textbf{Core} concepts directly correspond to the objects referred to in the query, and \textbf{broader} concepts, paired with the core concept, are aimed to capture more general categories that can help SAM3 detect instances missed by the core concept.
For example, given ``\textit{the person who stands up after sitting on the red couch}'', the core concepts would be \texttt{[person, couch]} and the paired broader concepts \texttt{[human, furniture]}, respectively.

\subsubsection{Mask Track Generation via SAM3.}
With the extracted concepts, we can directly infer SAM3 with the concepts as textual prompts to obtain mask tracks. Between the core and broader concepts from a pair in $C$, we select the one that yields more instances from SAM3 to prevent the pipeline from missing out objects, as illustrated in Fig.~\ref{fig:method}. We collect the mask tracks for each concept pair in $C$, yielding a set of mask tracks $M \in \{0, 1\}^{I \times T \times H \times W}$, which serve as a pool of \textbf{candidate masks}, where $I$ is the number of instances.

Nonetheless, even with core and broader concepts, SAM3 can still sometimes fail to detect an instance, \textit{i.e.} $I=0$, even with the broader concepts from the extraction stage. In this case, we cascade the extraction process while revising the concepts. Specifically, the subsequent iteration would generate more broadly scoped concepts for referring queries, or exploring alternative object categories for reasoning queries until it is identified by SAM3 in the video.

Each candidate mask $m_i \in M$ carries two types of information: i) spatial localization through the per-frame segmentation mask, and ii) temporal existence through the set of frames in which the object is present, denoted as $\mathcal{T}(m_i) = \{t \mid m_i^t \neq \emptyset\}$, where $m_i^t$ denotes the binary mask for instance $i$ at frame $t$.
This temporal existence information, a natural byproduct of SAM3's video-level processing, plays a central role in the subsequent pruning phase (Sec.~\ref{sec:method_pruning}).

% 3.4 Iterative Spatio-temporal Pruning
\subsection{Iterative Spatio-temporal Pruning}
\label{sec:method_pruning}
Given the generated pool of candidate masks $M$, we leverage the complex reasoning capabilities of MLLMs to obtain the query-grounded mask track $\mathcal{M}$, which the ability SAM3 lacks. As we have the spatio-temporal masks, we can naturally apply visual prompting~\cite{yang2023set, carion2025sam3} on the video to allow the MLLMs to further focus on the objects, as shown in Fig.~\ref{fig:method}.
However, evaluating all candidates in a single pass can be impractical: the pool may contain numerous objects with overlapping masks that would clutter the MLLM, and different candidates may appear at different temporal locations, making a single round of frame sampling insufficient to fairly assess all of them.
We therefore adopt an iterative pruning strategy that progressively reduces both the candidate set and the temporal scope, to allow the MLLM to concentrate its reasoning on increasingly fewer, harder cases.

\subsubsection{Candidate Pruning.}
At each iteration $r$, the MLLM examines the current candidate set $M^{(r)}$ and classifies each candidate into one of three categories:
\texttt{Accepted} (confidently matching the query), \texttt{Rejected} (confidently not matching), or \texttt{Uncertain} (requiring further evidence).
Accepted candidates are collected and merged into the output set 
$\mathcal{M} = \bigcup_r \{ m_i \in M^{(r)} \mid \texttt{Accepted} \}$; rejected candidates are permanently discarded.
Only uncertain candidates carry forward to the next iteration:
\begin{equation}
    M^{(r+1)} = \{m_i \in M^{(r)} \mid \texttt{Uncertain}\}.
\end{equation}
By committing to confident decisions early, the MLLM avoids repeatedly re-evaluating clear cases and concentrates its reasoning on genuinely ambiguous candidates.

\subsubsection{Temporal Scope Pruning.}
As the candidate set shrinks, the relevant temporal scope naturally contracts as well.
At iteration $r$, the system restricts the temporal scope to the union of frames where the remaining uncertain candidates exist:
\begin{equation}
    \mathcal{T}^{(r+1)} = \bigcup_{m_i \in M^{(r+1)}} \mathcal{T}(m_i).
\end{equation}
Frames are then sampled exclusively within $\mathcal{T}^{(r+1)}$, ensuring that every sampled frame contains at least one uncertain candidate. 

This achieves two effects simultaneously: it increases the density of informative content in the MLLM's visual input, and it reduces the total temporal span under consideration.
Notably, this mechanism requires no explicit frame selection module--the temporal narrowing emerges naturally from SAM3's temporal existence information.

\subsubsection{Convergence.}
The iterative process terminates when no uncertain candidates remain, \ie, $M^{(r+1)} = \emptyset$, meaning all candidates have been either accepted or rejected.
In practice, we also impose a maximum iteration count to bound computational cost.
Since $|M^{(r+1)}| \leq |M^{(r)}|$ at every non-trivial iteration, the process is guaranteed to terminate.
We observe that most queries converge within a small number of iterations, as the progressive narrowing rapidly reduces ambiguity (see Sec.~\ref{sec:experiments}).

%% file: sec/4_experiments.tex
\section{Experiments}
\label{sec:experiments}

\subsubsection{Datasets and Metrics.}
We evaluate our method on three major benchmarks for language-guided video object segmentation: MeViS~\cite{ding2023mevis}, ReVOS~\cite{yan2024visa}, and ReasonVOS~\cite{bai2024videolisa}. These datasets pose distinct challenges. MeViS features complex scenes involving multiple visually similar objects and demands strong motion understanding. ReVOS and ReasonVOS, on the other hand, emphasize reasoning-centric scenarios that require deeper semantic reasoning and world knowledge. Following prior works~\cite{kao2025cotrvs, bai2024videolisa}, we report region similarity \( \mathcal{J} \) (average IoU), contour accuracy \( \mathcal{F} \) (mean boundary similarity), and their average \( \mathcal{J} \& \mathcal{F} \).

\vspace{10pt}

\subsubsection{Implementation Details.}
We adopt various models as our baseline MLLMs: Qwen3-VL-8B-Thinking~\cite{bai2025qwen3} and Qwen3-VL-32B-Thinking for open-sourced models, and GPT-5~\cite{openai2025gpt5} for closed-source model.
As mentioned above, we use additional SAM3~\cite{carion2025sam3} to generate candidate mask tracks.
For each video, 16 frames are used by default. The maximum number of iterations is 3 for both concept extraction and iterative spatio-temporal pruning.
All experiments are conducted with 4 RTX PRO 5000 Blackwell GPUs.
Additional implementation details, including the prompts used in our pipeline, are provided in Appendix~\ref{sec:suppl_implementation}.

\input{tab/main_quan}

\vspace{10pt}

\subsection{Quantitative Results}
Tab.~\ref{tab:main_quan_referring_vos} presents quantitative comparisons with state-of-the-art methods on three challenging RVOS benchmarks: MeViS, ReVOS, and ReasonVOS. As shown in the table, \ours significantly outperforms existing training-free methods achieving up to \textbf{40.0\%}, \textbf{18.6\%}, \textbf{15.3\%} improvements on MeViS, ReVOS, and ReasonVOS respectively. Even when using the same backbone model, \ours significantly outperforms prior approaches. For instance, with Qwen3-VL-8B-Thinking, \ours surpasses CoT-RVS~\cite{kao2025cotrvs} using the same model. 
Moreover, even when compared with pipelines that integrate powerful closed-source models, such as AL-Ref-SAM2~\cite{huang2025alrefsam2} and CoT-RVS with GPT-4o~\cite{hurst2024gpt}, our method still achieves substantially better performance, highlighting the effectiveness of our pipeline design.
These results suggest that the spatio-temporal information provided by SAM3 effectively supports MLLMs in both temporal understanding and reasoning over complex video queries.

Furthermore, when stronger MLLMs are integrated into our framework, the performance consistently improves. Using Qwen3-VL-32B-Thinking, \ours achieves state-of-the-art performance across all benchmarks, even compared with training based approaches. Replacing the backbone with strong closed-source models like GPT-5 further improves the results, demonstrating the scalability of our pipeline leveraging stronger MLLM reasoning capabilities.

\input{fig/qualitative}
\subsection{Qualitative Results}
Fig.~\ref{fig:qualitative} presents qualitative comparisons between \ours and CoT-RVS on several challenging referring expressions. In the first example, where multiple instances of the same category appear, \ours correctly identifies the target by reasoning over inter-object relations, while CoT-RVS fails to distinguish between similar objects. In the second example, our method successfully resolves temporal motion reasoning by identifying the vehicle moving to the left among multiple vehicles. In the third example, \ours accurately performs instance-level reasoning, distinguishing the correct lamb among the same instances. These results demonstrate the effectiveness of \ours in accurately segmenting the correct objects under challenging scenarios by enabling stronger query-grounded reasoning. Additional overall reasoning process and qualitative results are provided in the Appendices~\ref{sec:suppl_overall_reasoning} and \ref{sec:suppl_qual}, respectively.

\subsection{Analysis}
\input{tab/ablation}

\subsubsection{Component analysis.} In Tab.~\ref{tab:abl_pipeline}, we provide ablation studies for our core components--(a) retry for the concept extraction in Candidate Mask Track Generation, (b) iteration and (c) temporal scope pruning in Iterative Spatio-temporal Pruning. 
For concept extraction, we can clearly observe that the (a) retry strategy, play a significant role for preventing the framework from falling into failure cases where SAM3 is not able to detect any objects. We provide detailed analysis of concept extraction in Tab.~\ref{tab:empty_ratio} and in Fig.~\ref{fig:noun_extraction_iter} below.

For the Iterative Spatio-temporal Pruning phase, we can also observe that both the (b) iteration and (c) temporal scope pruning play a significant role, evident from the gains for our full pipeline. This verifies the effectiveness of the core motivation for our framework - MLLMs can perform better reasoning when the model is able to spatio-temporally \textbf{focus} in a video, as opposed to existing approaches that na\"ively employ the MLLMs to reason the entire video at once, with sampled frames. Additional analysis on the effect of MLLM reasoning is provided in Appendix~\ref{sec:suppl_sam3_ablation}.

% ================= Table 3: Max Iteration =================
\begin{wraptable}{r}{0.4\columnwidth}
\centering
\renewcommand{\arraystretch}{0.96} \resizebox{\linewidth}{!}{ \begin{tabular}{L{2.5cm}|Z{1.0cm}Z{1.0cm}Z{1.2cm}}
\toprule
Max Iteration & \( \mathcal{J} \) & \( \mathcal{F} \) & \( \mathcal{J}\&\mathcal{F} \) \\
\midrule
1 (No Retry) & 63.0 & 68.1 & 65.6 \\
2 & 64.4 & 70.6 & 67.5 \\
\rowcolor{blue}
3 & 67.1 & 73.6 & 70.3 \\
4 & 67.4 & 73.7 & 70.6 \\
\bottomrule
\end{tabular}%
}
\caption{\textbf{Results for varying maximum number of iterations.}}
\vspace{-15pt}
\label{tab:abl_iter}
\end{wraptable}

\subsubsection{Ablation on maximum number of iterations for spatio-temporal pruning.}
In Tab.~\ref{tab:abl_iter}, we provide quantitative results for varying number of max iterations we apply for Iterative Spatio-temporal Pruning. As shown, we can clearly observe that the iterations show steady improvements, with significant gains up to 3 iterations. As discussed in Sec.~\ref{sec:method_pruning}, we observe that most queries converge in small number of iterations, being 3 in particular, and show minimal improvements with additional iterations further on. Therefore, we set our maximum number of iterations as 3, further iterating the process would increase the inference time with marginal improvements.

\subsubsection{Ablation on number of sampled frames.}

\begin{wraptable}{r}{0.4\columnwidth}
\vspace{-22pt}
\centering
\renewcommand{\arraystretch}{0.96}
\resizebox{0.4\columnwidth}{!}{%
\begin{tabular}{L{2.5cm}|Z{1.2cm}Z{1.2cm}Z{1.4cm}}
\toprule
\# of Frames & \( \mathcal{J} \) & \( \mathcal{F} \) & \( \mathcal{J}\&\mathcal{F} \) \\
\midrule
8 & 64.6 & 70.7 & 67.6 \\
\rowcolor{blue}
16 & 67.1 & 73.6 & 70.3 \\
32 & 65.6 & 72.1 & 68.8 \\
64 & 64.9 & 71.5 & 68.2 \\
\bottomrule
\end{tabular}%
}
\caption{\textbf{Effect of the number of sampled frames.}}
\label{tab:abl_frames}
\end{wraptable}

Our empirical results show that sampling 16 frames yields the best performance of 70.3 \( \mathcal{J} \& \mathcal{F} \). Using fewer frames (8 in this table) limits temporal coverage, while increasing beyond 16 introduces redundancy that may hinder the reasoning ability of the MLLM, with diminishing returns in both accuracy and computational cost. We use 16 frames as the default for all other experiments.

\subsubsection{Effectiveness of iteration for Iterative Spatio-temporal Pruning.}
\input{fig/pruning_iter}

In Fig.~\ref{fig:pruning_iter}, we provide detailed visual analysis of the effects of iteration for our Iterative Spatio-temporal Pruning phase. For simple cases, as shown on the top, we can observe that the MLLM is able to confidently classify objects into \texttt{Accepted} or \texttt{Rejected}, thus not resulting in redundant iterations. In more challenging examples, we can observe that the MLLM is capable of identifying objects that the model is uncertain at first glance. As for the example shown in the middle bottom row, we can observe the framework iteratively rejecting objects and focusing on more confusing objects.

\subsubsection{Effectiveness of iteration for concept extraction.}
\input{fig/noun_extraction_iter}

\begin{wraptable}{r}{0.4\columnwidth}
\vspace{-22pt}
\centering
\renewcommand{\arraystretch}{0.96}
\resizebox{\linewidth}{!}{%
\begin{tabular}{L{2.5cm}|Z{4.5cm}}
\toprule
Iterations & Empty Mask Ratio (\%) \\
% \cmidrule(lr){2-4}
\midrule
CoT-RVS~\cite{kao2025cotrvs} & 12.0 \\
\midrule
1 & 4.9 \\
2 & 4.1 (\textcolor{ForestGreen}{-0.8})\\
\rowcolor{blue} 3 & 3.8 (\textcolor{ForestGreen}{-1.1}) \\
\bottomrule
\end{tabular}%
}

\caption{\textbf{Effects of iterations on the ratio of empty mask prediction (\%).}} 
\label{tab:empty_ratio}
\vspace{-15pt}
\end{wraptable}

In Tab.~\ref{tab:empty_ratio}, we report the ratio of empty masks generated from the Candidate Mask Track Generation phase for different number of iterations applied for the concept extraction process. As a comparison, we also report the ratio of empty masks from CoT-RVS~\cite{kao2025cotrvs}, where CoT-RVS leverages CLIP to identify query-relevant frames to reason with the MLLM. We can clearly observe that even without additional iterations, we show significantly lower ratio for having empty masks, demonstrating the effectiveness of our approach with MLLM and SAM3. Further adding iterations to the concept extraction stage minimizes the ratio for having empty masks generated from SAM3.

We further provide detailed visual analysis in Fig.~\ref{fig:noun_extraction_iter}, where in Fig.~\ref{fig:noun_extraction_iter}(a) the framework is able to identify objects that were not previously identified for the \textbf{reasoning} queries (\eg \texttt{folded paper}). Furthermore, we can also observe cases where the subsequent retry processes revise the concept as a more general (Fig.~\ref{fig:noun_extraction_iter}(b)) or more specific (Fig.~\ref{fig:noun_extraction_iter}(c)) concept for the \textbf{referring} queries, allowing SAM3 to correctly identify objects that it was previously not able to segment.

%% file: tab/main_quan.tex
\begin{table}[t]
\centering
\renewcommand{\arraystretch}{0.96}
\caption{\textbf{Comparison with state-of-the-art methods on Referring and Reasoning VOS Benchmarks}: MeViS~\cite{ding2023mevis}, ReVOS~\cite{yan2024visa} and ReasonVOS~\cite{bai2024videolisa}. Qwen3-VL-8B-T and Qwen3-VL-32B-T indicate Qwen3-VL-8B-Thinking~\cite{bai2025qwen3} and Qwen3-VL-32B-Thinking model, respectively. The best performing results are presented in \textbf{bold}, while the second-best results are \underline{underlined}. $\dagger$ denotes our reproduced results.}
\label{tab:main_quan_referring_vos}

\resizebox{\textwidth}{!}{%
\begin{tabular}
{L{3.6cm}Z{2.8cm}
 Z{0.85cm}Z{0.85cm}Z{1.05cm}
 Z{0.85cm}Z{0.85cm}Z{1.05cm}
 Z{0.85cm}Z{0.85cm}Z{1.05cm}
 Z{0.85cm}Z{0.85cm}Z{1.05cm}
 Z{0.85cm}Z{0.85cm}Z{1.05cm}}
\toprule
\multirow{2}{*}[-10pt]{Method} &
\multirow{2}{*}[-10pt]{MLLM} &
\multicolumn{3}{c}{MeViS} &
\multicolumn{9}{c}{ReVOS} &
\multicolumn{3}{c}{ReasonVOS} \\
\cmidrule(lr){3-5}
\cmidrule(lr){6-14}
\cmidrule(lr){15-17}
 & &
\multirow{2}{*}{\( \mathcal{J} \)} & \multirow{2}{*}{\( \mathcal{F} \)} & \multirow{2}{*}{\( \mathcal{J}\&\mathcal{F} \)} &
\multicolumn{3}{c}{Referring} &
\multicolumn{3}{c}{Reasoning} &
\multicolumn{3}{c}{Overall} &
\multirow{2}{*}{\( \mathcal{J} \)} & \multirow{2}{*}{\( \mathcal{F} \)} & \multirow{2}{*}{\( \mathcal{J}\&\mathcal{F} \)} \\
\cmidrule(lr){6-8}
\cmidrule(lr){9-11}
\cmidrule(lr){12-14}
 & &
 & & &
{\( \mathcal{J} \)} & \( \mathcal{F} \) & \( \mathcal{J}\&\mathcal{F} \) &
\( \mathcal{J} \) & \( \mathcal{F} \) & \( \mathcal{J}\&\mathcal{F} \) &
\( \mathcal{J} \) & \( \mathcal{F} \) & \( \mathcal{J}\&\mathcal{F} \) &
 & & \\
\midrule
\midrule
\multicolumn{17}{l}{\textit{Training-based Methods}} \\
\midrule
VideoLISA~\cite{bai2024videolisa} & LLaVA-3.8B
& 41.3 & 47.6 & 44.4
& - & - & - & - & - & - & - & - & - & 45.1 & 49.9 & 47.5 \\
VISA~\cite{yan2024visa} & Chat-UniVi-13B
& 41.8 & 47.1 & 44.5
& 55.6 & 59.1 & 57.4 & 42.0 & 46.7 & 44.3 & 48.8 & 52.9 & 50.9
& - & - & - \\
HyperSeg~\cite{wei2024hyperseg} & Mipha-3B
& - & - & -
& 56.0 & 60.9 & 58.5 & 50.2 & 55.8 & 53.0 & 53.1 & 58.4 & 55.7
& - & - & - \\
InstructSeg~\cite{wei2025instructseg} & Mipha-3B
& - & - & -
& 54.8 & 59.2 & 57.0 & 49.2 & 54.7 & 51.9 & 52.0 & 56.9 & 54.5
& - & - & - \\
GLUS~\cite{lin2025glus} & LISA-7B
& 48.5 & 54.2 & 51.3
& 56.0 & 60.7 & 58.3 & 48.8 & 53.9 & 51.4 & 52.4 & 57.3 & 54.9
& 47.5 & 52.4 & 49.9 \\
ViLLa~\cite{varma2023villa} & InternVideo2-6B
& 46.5 & 52.3 & 49.4
& - & - & - & - & - & - & 54.9 & 59.1 & 57.0
& - & - & - \\
Sa2VA~\cite{yuan2025sa2va} & InternVL2-8B
& - & - & 46.9
& - & - & - & - & - & - & - & - & 57.6
& - & - & - \\
Sa2VA~\cite{yuan2025sa2va} & InternVL3-14B
& - & - & -
& - & - & - & - & - & - & - & - & 60.7
& - & - & - \\
RGA3~\cite{wang2025object} & Qwen2.5-VL-7B
& 47.4 & 52.8 & 50.1
& 58.7 & 62.3 & 60.5 & 53.1 & 57.7 & 55.4 & 55.9 & 60.0 & 58.0
& 51.3 & 56.0 & 53.6 \\
VRS-HQ~\cite{gong2025devil} & Chat-UniVi-13B
& 48.0 & 53.7 & 50.9
& 61.1 & 65.5 & 63.3 & 54.1 & 59.4 & 56.8 & 57.6 & 62.5 & 60.0
& - & - & - \\
VideoSeg-R1~\cite{xu2025videosegr1} & Qwen2.5-VL-7B
& 52.7 & 57.8 & 55.3
& - & - & - & - & - & - & 58.2 & 64.0 & 61.1
& - & - & - \\
\midrule
\multicolumn{17}{l}{\textit{Training-free Methods}} \\
\midrule
AL-Ref-SAM2~\cite{huang2025alrefsam2} & GPT-4
& 39.5 & 46.2 & 42.8
& - & - & - & - & - & - & - & - & -
& - & - & - \\
CoT-RVS~\cite{kao2025cotrvs} & Gemma3-12B
& 40.3 & 48.1 & 44.2
& - & - & - & - & - & - & 43.4 & 50.9 & 47.1
& 47.5 & 54.0 & 50.7 \\
$\text{CoT-RVS}^\dagger$~\cite{kao2025cotrvs} & Qwen3-VL-8B-T
& 37.7 & 43.9 & 40.8
& 56.1 & 61.5 & 58.8 & 46.6 & 53.0 & 49.8 & 51.4 & 57.3 & 54.3
& 52.5 & 58.9 & 55.7 \\
CoT-RVS~\cite{kao2025cotrvs} & GPT-4o
& 48.7 & 55.7 & 52.2
& - & - & - & - & - & - & 52.8 & 59.0 & 55.9
& 62.4 & 68.7 & 65.5 \\
\midrule
\rowcolor{blue}
 & Qwen3-VL-8B-T
& 59.2 & 64.5 & 61.9
& 58.7 & 62.5 & 60.6 & 56.8 & 61.3 & 59.0 & 57.7 & 61.9 & 59.8
& 65.5 & 71.8 & 68.6 \\
\rowcolor{blue}
 & Qwen3-VL-32B-T
& \underline{65.3} & \underline{70.0} & \underline{67.7}
& \underline{62.8} & \underline{66.8} & \underline{64.8} & \underline{58.0} & \underline{62.6} & \underline{60.3} & \underline{60.4} & \underline{64.7} & \underline{62.5}
& \underline{67.3} & \underline{73.4} & \underline{70.4} \\
\rowcolor{blue}
\multirow{-3}{*}{\textbf{\ours} (Ours)} & GPT-5
& \textbf{70.4} & \textbf{75.7} & \textbf{73.1}
& \textbf{66.8} & \textbf{70.8} & \textbf{68.8} & \textbf{61.4} & \textbf{66.0} & \textbf{63.7} & \textbf{64.1} & \textbf{68.4} & \textbf{66.3}
& \textbf{73.1} & \textbf{78.0} & \textbf{75.5} \\
\bottomrule
\end{tabular}%
}

\end{table}

%% file: fig/qualitative.tex
\begin{figure*}[t]
    \centering
    \includegraphics[width=\textwidth]{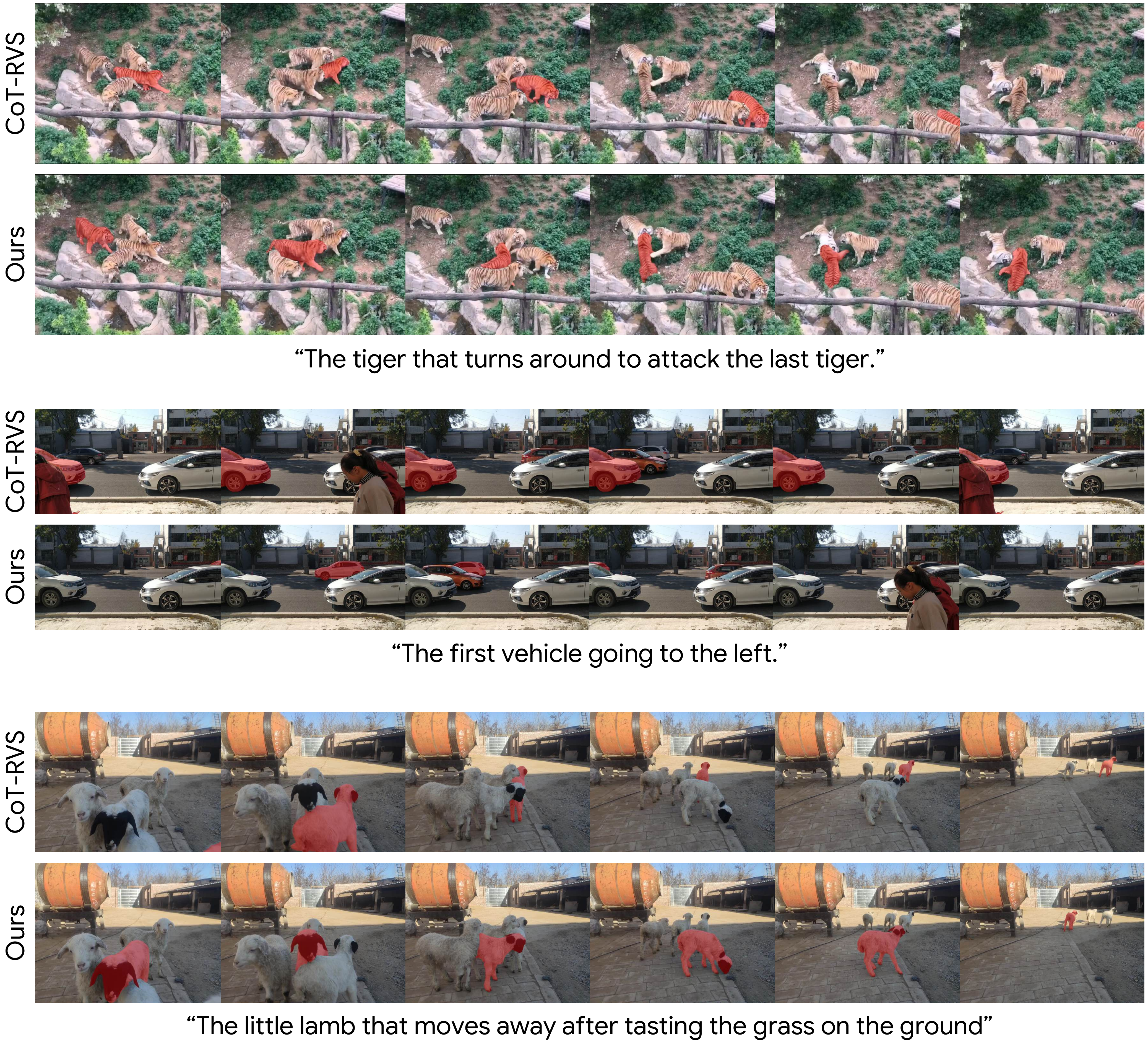}
    \caption{\textbf{Qualitative results.}
    \ours effectively resolves challenging scenarios such as multi-instance ambiguity and temporal reasoning, accurately segmenting the referred objects.
    }
    \label{fig:qualitative}
\end{figure*}

%% file: tab/ablation.tex
% ================= Table 1: Pipeline =================
\begin{table}[t]
\centering
\renewcommand{\arraystretch}{0.96}

\caption{\textbf{Component analysis.} We provide ablation study for our core components--\textbf{(a)} retry for the concept extraction in the Candidate Mask Track Generation phase, and \textbf{(b)} the iterative process and \textbf{(c)} temporal scope pruning in the Iterative Spatio-temporal Pruning phase.}
\label{tab:abl_pipeline}

\resizebox{0.9\textwidth}{!}{%
\begin{tabular}{L{9cm}|Z{1.0cm}Z{1.0cm}Z{1.2cm}}
\toprule
Component & \( \mathcal{J} \) & \( \mathcal{F} \) & \( \mathcal{J}\&\mathcal{F} \) \\
\midrule
\textbf{(a)} w/o Retry in Concept Extraction & 65.6 & 71.8 & 68.7 \\
\textbf{(b)} w/o Iteration in Iterative Spatio-temporal Pruning & 63.0 & 68.1 & 65.6 \\
\textbf{(c)} w/o Temporal Scope Pruning & 64.2 & 70.4 & 67.3 \\
\midrule
\rowcolor{blue}
\textbf{\ours} (Full Pipeline) & 67.1 & 73.6 & 70.3 \\
\bottomrule
\end{tabular}%
}
\vspace{-10pt}
\end{table}

%% file: fig/pruning_iter.tex
\begin{figure*}[t]
    \centering
    \includegraphics[width=\textwidth]{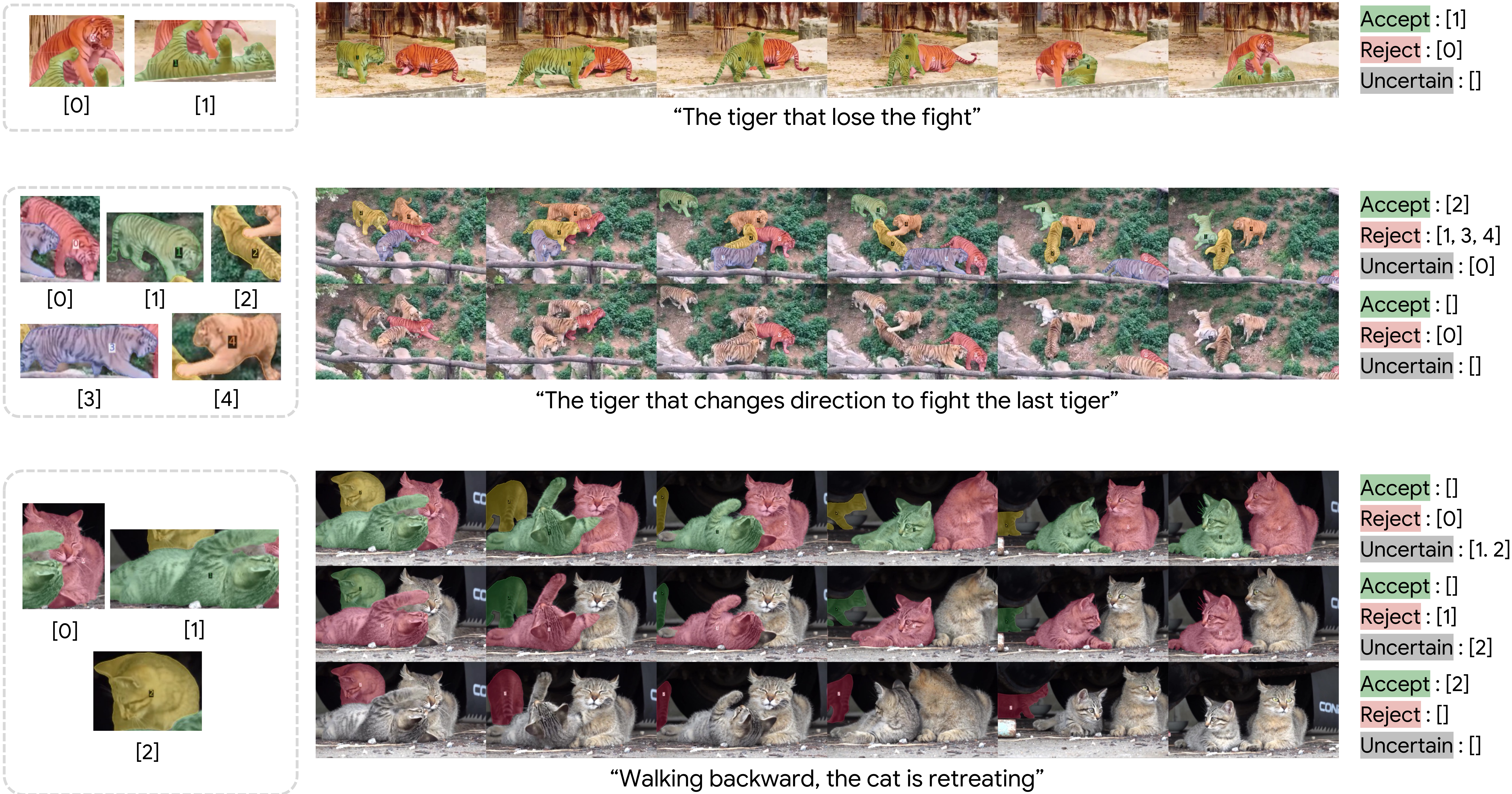}
    \caption{\textbf{Qualitative results of iteration in Iterative Spatio-temporal Pruning.}
    We illustrate how our iterative spatio-temporal pruning progressively narrows the relevant temporal window and eliminates irrelevant track candidates.
    Across iterations, the remaining candidates become fewer but more query-consistent, leading to the final selected track set.}
    % \vspace{-10pt}
    \label{fig:pruning_iter}
\end{figure*}

%% file: fig/noun_extraction_iter.tex
\begin{figure*}[!h]
    \centering
    \includegraphics[width=\textwidth]{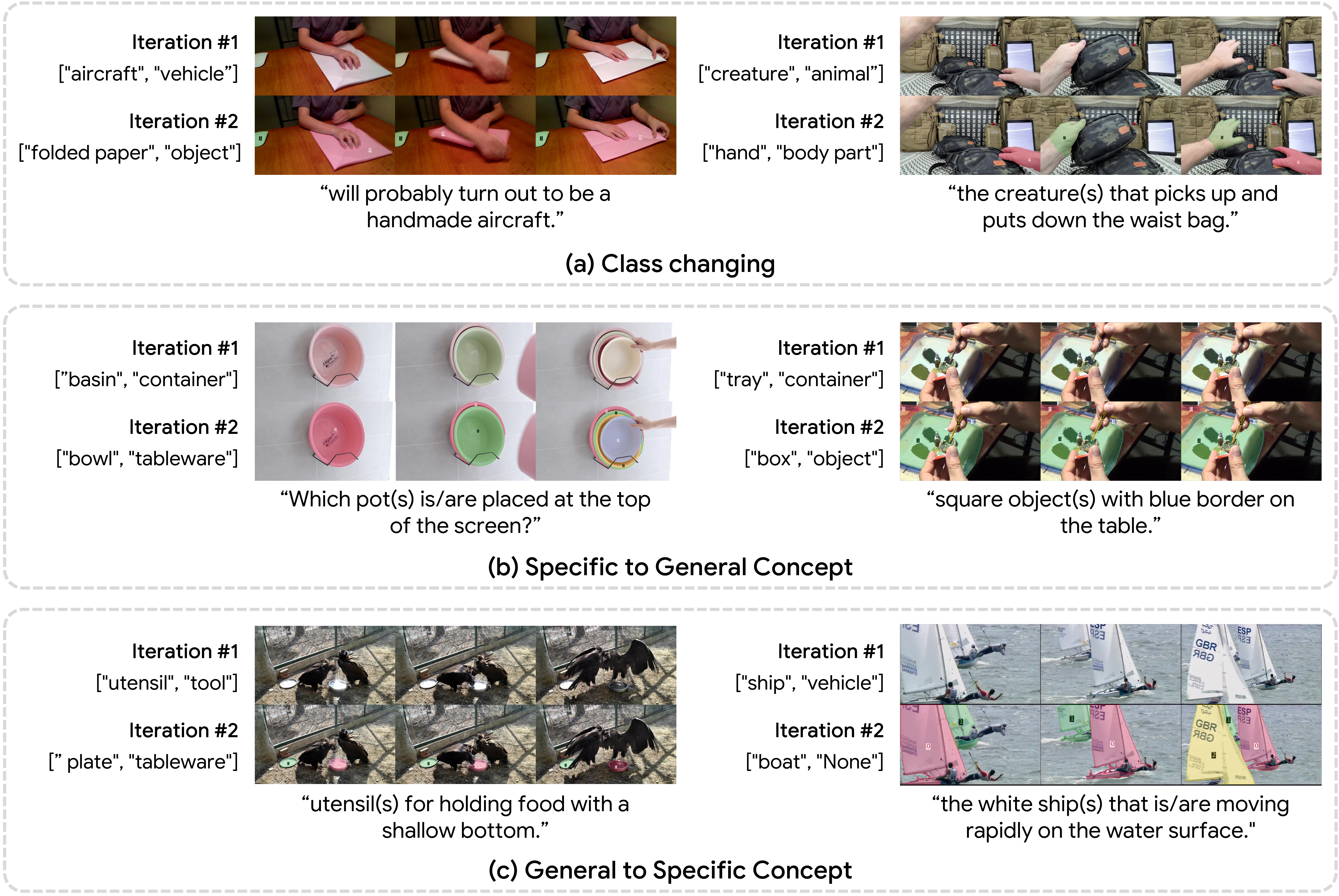}
    \caption{\textbf{Qualitative results of iteration in Concept Extraction.}
    We visualize how our iterative concept extraction progressively refines object concepts from the query. From top to bottom, the iterative process refines concepts through three distinct patterns: (a) the extracted concept class changes entirely to better match the query, (b) a specific concept is broadened to a more general one, and (c) a vague concept is narrowed to a more precise one, each converging to concepts better aligned with the query's intent.
    }
    \label{fig:noun_extraction_iter}
    % \vspace{-15pt}
\end{figure*}

%% file: sec/5_conclusion.tex
\section{Conclusion}
In this paper, we present \ours, a training-free agentic pipeline that combines SAM3's language-grounded perception with the MLLM's reasoning capability for Referring Video Object Segmentation tasks. By leveraging SAM3 for object detection and temporal localization, our pipeline enables the MLLM to perform query-grounded reasoning over structured object-level evidence rather than the entire video directly.
To effectively utilize this complementary design, we introduce an iterative spatio-temporal pruning strategy that progressively narrows the candidate set and temporal scope, allowing the MLLM to focus on ambiguous objects and refine its reasoning iteratively.
Extensive experiments across multiple RVOS benchmarks and various MLLM backbones demonstrate the effectiveness of \ours, highlighting the complementary strengths of SAM3’s spatio-temporal perception and MLLM-based reasoning.

%% file: sec/suppl.tex
\appendix

\setcounter{secnumdepth}{2}
\setcounter{section}{0}
\setcounter{figure}{0}
\setcounter{table}{0}

\renewcommand{\thefigure}{A\arabic{figure}}
\renewcommand{\thetable}{A\arabic{table}}

%%%%%%%%%%%%%%%%%%%%%%%%%%%%%%%%%%%%%%%%%%%%%%%%%%%%%%%%%%%%%%%%%%%%%%%%%%%%%%%%%%%%%%%%%%%%%%%%
% Overview
\section*{Appendix Overview}
The appendix is organized as follows.

\begin{itemize}
    \item \textbf{Sec.~\ref{sec:suppl_results}} presents additional experiments and analyses: \\
        - Ablation on the effect of MLLM reasoning (Sec.~\ref{sec:suppl_sam3_ablation}) \\
        - Robustness of Concept Extraction (Sec.~\ref{sec:suppl_concept_extraction}) \\
        - Details of Visual Prompting (Sec.~\ref{sec:suppl_appearance}) \\
        - Detailed Algorithm of \ours (Sec.~\ref{sec:suppl_algorithm}) \\
        - Visualization of Overall Reasoning Process of \ours (Sec.~\ref{sec:suppl_overall_reasoning}) \\
        - Additional Qualitative Results (Sec.~\ref{sec:suppl_qual}) \\
    \item \textbf{Sec.~\ref{sec:suppl_details}} provides implementation details of \ours: \\
        - Ablation study configurations (Sec.~\ref{sec:suppl_implementation}) \\
        - Detailed prompts used in \ours (Sec.~\ref{sec:suppl_prompts}) \\
    \item \textbf{Sec.~\ref{sec:suppl_future_works}} discusses future directions. \\
\end{itemize}

\section{Additional Results and Analyses}
\label{sec:suppl_results}

\subsection{Ablation on the Effect of MLLM reasoning}
\label{sec:suppl_sam3_ablation}
\vspace{-10pt}
\input{tab/suppl_sam3_ablation}
\vspace{-5pt}
We evaluate the necessity of MLLM-based reasoning in our agentic pipeline by replacing it with direct input of the referring expression into SAM3~\cite{carion2025sam3}.
Since SAM3 is designed to process simple noun phrases rather than complex linguistic queries, this baseline bypasses query-grounded reasoning entirely.
As shown in Tab.~\ref{tab:suppl_sam3_ablation}, removing MLLM reasoning leads to a substantial performance drop of 29.5 points in $\mathcal{J}\&\mathcal{F}$ (40.8 vs.\ 70.3), alongside a sharp increase in the empty mask ratio from 3.8\% to 38.9\%, confirming that SAM3 alone cannot adequately handle the semantic complexity of referring expressions. These results demonstrate that the complementary roles of SAM3 and the MLLM are both necessary, where SAM3 provides reliable perceptual grounding while MLLM reasoning is essential for interpreting the query and resolving ambiguity among candidate tracks.

\subsection{Robustness of Concept Extraction}
\label{sec:suppl_concept_extraction}
\input{fig/suppl_concept_extraction}

We provide further detail on the concept extraction step introduced in Sec. 3, focusing on how it handles diverse query types and why it reliably satisfies the coverage requirement of the candidate generation phase.
The behavior of concept extraction is driven by the type of the given language query $Q$. Although this step may appear fragile--particularly for queries that resist direct noun extraction (\eg ``\textit{what made the dogs move}'')--our pipeline explicitly accounts for the query type before invoking SAM3.

Specifically, we distinguish between two cases.
For \textbf{referring} queries, the target object is identifiable from $Q$ alone without visual evidence (Fig.~\ref{fig:suppl_concept_extraction} (a)). In this case, concept extraction proceeds from text only, and the resulting concepts reliably anchor SAM3 to the correct object category.
For \textbf{reasoning} queries, the target object cannot be determined without observing the video (Fig.~\ref{fig:suppl_concept_extraction} (b)). Here, the MLLM examines uniformly sampled frames alongside $Q$ to infer plausible candidate concepts. In most cases, this is sufficient to identify the target object. However, in extreme cases the referred object may not be clearly visible in the sampled subset--a hand appearing in only two frames, for instance, could be entirely absent from a uniform sample. Even so, the MLLM can still propose plausible candidates such as ``\textit{person}'' or ``\textit{hand}'' by reasoning about the expression and the available visual context. Because SAM3 processes all frames rather than only the sampled subset, it can verify whether these candidates actually exist anywhere in the video, reliably capturing objects that uniform sampling alone would miss.

This design directly addresses the coverage requirement identified in Sec. 3: candidate tracks that do not contain the referred object cannot be recovered by subsequent reasoning.
By converting the language query into SAM3-compatible noun phrases according to the query type, concept extraction ensures that the candidate pool is sufficiently complete before the spatio-temporal pruning phase begins.

% Details about visual prompting
\subsection{Details of Visual Prompting}
\label{sec:suppl_appearance}
\input{fig/suppl_appearance}
In the iterative spatio-temporal pruning stage, the MLLM reasons over mask-overlaid videos where each candidate mask track is visualized with a distinct color~\cite{yang2023set, carion2025sam3}. While this visual prompting strategy is essential for grounding the MLLM's focus to specific object regions, it inevitably obscures the original appearance of each instance, including color, texture, and other fine-grained visual details. This can be problematic when the referring expression hinges on such appearance cues (\eg, ``\textit{Parking white car}'').

To mitigate this, we introduce a simple appearance tool that can be optionally invoked before pruning begins. A brief illustration is provided in Fig.~\ref{fig:suppl_appearance}. When the MLLM determines that appearance-level evidence is needed, the tool constructs a two-panel image (which we refer to as a \textbf{reference image}) for each candidate following a similar visualization strategy to SAM3 Agent~\cite{carion2025sam3}: a loosely cropped view with the bounding box for spatial context, and a tightly cropped view for fine-grained appearance.
Notably, for each candidate object, we select the single frame index where the object has the largest visible area, which is obtained from the mask track information produced by SAM3.
The MLLM then generates a brief natural language description of each candidate's appearance, which is carried forward into the pruning stage as supplementary evidence. Additionally, as shown in Tab.~\ref{tab:suppl_appearance} (the results were obtained under the same setting as the ablation studies), incorporating the appearance tool yields some improvement, confirming that it effectively recovers appearance information lost under the visual overlay.

\input{tab/suppl_appearance}

% Algorithm
\subsection{Detailed Algorithm}
\label{sec:suppl_algorithm}
We provide a detailed description of the \ours pipeline through three algorithms. Algorithm~\ref{alg_concept_extraction} (Candidate Mask Track Generation) generates candidate mask tracks by extracting concept pairs from the query using an MLLM and prompting SAM3 with these concepts. Algorithm~\ref{alg_appearance_tool} (Appearance Tool) introduces an appearance tool that determines whether additional appearance information is required by the query and extracts it when necessary. Algorithm~\ref{alg_selection} (Iterative Spatio-temporal Pruning) performs iterative spatio-temporal pruning, where candidate mask tracks are progressively verified and filtered through reasoning over the video and the query.

\subsubsection{Candidate Mask Track Generation.}
Algorithm~\ref{alg_concept_extraction} describes the procedure for generating mask tracks from a natural language query. Given a video $\mathcal{V}=\{I_t\}^T_{t=1}$ and a language query $Q$, the algorithm iteratively extracts concept pairs using an MLLM and uses them to prompt SAM3 for mask track generation.

At the first iteration ($k=0$), the MLLM receives $Q$ with $\texttt{prompt}_{referring}$(\S~\ref{fig:suppl_prompts_1}). The model first determines the query type. If the query corresponds to a \textit{referring} case, where the referred object can be identified from the textual expression alone, the model directly extracts a set of concept pairs from the query. Otherwise, if the query is classified as \textit{reasoning} and requires visual information from the video, the model does not extract concept pairs at this stage and instead proceeds to perform video conditioned concept extraction using $\texttt{prompt}_{reasoning}$(\S~\ref{fig:suppl_prompts_2}). Each extracted concept pair, indexed by $i$, consists of a core concept $c^{core}_i$ that closely corresponds to the referred object and a broader concept $c^{broad}_i$ that represents a more general category, increasing the likelihood of retrieving candidate tracks. For each concept pair ($c^{core}_i,c^{broad}_i$), SAM3 is applied to the video separately using the core concept and the broader concept, producing two candidate mask tracks.
Among the concepts, we retain the track with the larger number of the sum of the total detected instances across frames, and denote this number as $\textsc{Count}(\cdot)$.
If no instances are detected, a retry step is performed. Previously failed concept pairs are accumulated in a failure set and provided to the MLLM in the next iteration to prevent the model from generating similar concepts again. This process continues until a mask track containing at least one detected instance is obtained or the maximum retry limit $k_{max}$, which is set to 3, is reached. The result of this stage is a set of candidate mask tracks $M$ together with the final selected concepts $C^*$ which provides class-level information about the target object for the subsequent selection stage.

\subsubsection{Appearance Tool.}
Since mask overlays used during candidate visualization may obscure appearance cues such as color, an additional appearance tool is introduced to explicitly extract appearance information when needed. Algorithm~\ref{alg_appearance_tool} introduces an appearance tool that extracts additional appearance information when required by the query (\eg ``\textit{a white dog with black dots}''). The MLLM first analyzes the query using $\texttt{prompt}_{appearance\_requirement}$(\S~\ref{fig:suppl_prompts_appearance} (top)) to determine whether appearance attributes are necessary for identifying the target object.
If appearance information is required, the MLLM extracts appearance cues using $\texttt{prompt}_{appearance\_retrieval}$(\S~\ref{fig:suppl_prompts_appearance} (bottom)).
The MLLM then receives reference images $I$ for all candidate objects as described in Sec.~\ref{sec:suppl_appearance}, and extracts an appearance description $\mathcal{A}$ for all candidate objects.

\subsubsection{Iterative Spatio-temporal Pruning.}
Algorithm~\ref{alg_selection} performs the final selection of the referred object through iterative spatio-temporal pruning of candidate mask tracks. The procedure begins with the candidate mask tracks $M$ obtained from Algorithm~\ref{alg_concept_extraction}. The initial temporal scope $\mathcal{T}^{(0)}$ is defined as the union of frame indices where at least one instance is detected in the candidate mask tracks.

To proceed with the selection process, the candidate mask tracks are first visualized on the video frames, producing a masked overlaid video $V^*$. At each iteration $r$, frames are uniformly sampled from the current temporal scope $\mathcal{T}^{(r)}$. Each candidate mask track $m_i \in M^{(r)}$ is then evaluated by the MLLM using $\texttt{prompt}_{select}$(\S~\ref{fig:suppl_prompts_pruning}), conditioned on the mask overlaid video $V^*$, the query $Q$, the concept for each object $C^*$, and the appearance information $\mathcal{A}$.

Based on this evaluation, each mask track is classified into one of three categories: \texttt{Accepted}, \texttt{Rejected}, or \texttt{Uncertain}. Tracks classified as \texttt{Accepted} are added to the final mask set $\mathcal{M}$, while tracks labeled \texttt{Rejected} are removed from further consideration. Tracks labeled \texttt{Uncertain} are retained for further evaluation in the next iteration.

As the pruning process progresses, the temporal scope is updated using the temporal spans associated with the uncertain tracks. Specifically, the next temporal set $\mathcal{T}^{(r+1)}$ is defined as the union of the temporal intervals $\mathcal{T}(m_i)$ of the uncertain tracks. At the same time, visualization is also restricted to these uncertain candidates, which further reduces the spatial region under consideration. Together, these updates progressively narrow both the temporal and spatial search space to regions where ambiguous candidates remain.

The iterative pruning continues until no uncertain tracks remain or the maximum iteration limit $r_{max}$, which is set to 3, is reached. The final output of this stage is the set of predicted masks $\mathcal{M}$.

\input{tab/suppl_algorithm}

\FloatBarrier

% Overall reasoning pipeline
\subsection{Visualization of Overall Reasoning Process}
\label{sec:suppl_overall_reasoning}

We provide end-to-end qualitative visualizations of \ours in Figs.~\ref{fig:suppl_sample_1} and ~\ref{fig:suppl_sample_2}, illustrating how the pipeline processes a referring expression from concept extraction through iterative spatio-temporal pruning to the final mask track output.

In Fig.~\ref{fig:suppl_sample_1}, the referring expression is ``\textit{Which car disappears from the scene first?}'' The MLLM extracts the core concept ``\textit{car}'' and the broader concept ``\textit{vehicle}.'' Since SAM3 detects only 2 objects with the core concept, the pipeline falls back to the broader concept, yielding 4 candidate mask tracks. During iterative spatio-temporal pruning, the MLLM first rejects objects 2 and 3 as bicycles while classifying objects 0 and 1 as uncertain, since both cars remain visible in the sampled frames. In the subsequent iteration, with only the two cars remaining, the MLLM examines their temporal presence more carefully and determines that object 1 disappears first, accepting it as the final output. This example demonstrates a case where candidate pruning alone suffices to resolve the expression.

Fig.~\ref{fig:suppl_sample_2} presents a more complex case with the expression ``\textit{Which creature has the minimum energy loss?}'' Again, the core concept ``\textit{monkey}'' retrieves only 2 objects, so the pipeline falls back to ``\textit{animal},'' yielding 4 candidates. The first pruning iteration rejects the clearly active zebras (objects 2 and 3) while marking objects 4 and 5 as uncertain. In the next iteration, the MLLM observes that object 5 moves more noticeably and rejects it, but remains uncertain about object 4. At this point, temporal scope pruning is triggered: since object 4 mostly does not span the full temporal range of the video, the pipeline restricts reasoning to the temporal scope where only object 4 exists. With this pruned context, the MLLM confirms that object 4 moves less actively compared to the zebras and accepts it as the final answer. This example shows how temporal scope pruning complements candidate pruning when spatial reasoning alone cannot fully resolve the expression.

\input{fig/suppl_overall_reasoning}

\subsection{Additional Qualitative Results}

We present additional qualitative results of \ours on MeViS~\cite{ding2023mevis} (Figs.~\ref{fig:suppl_qual_mevis_1},~\ref{fig:suppl_qual_mevis_2}), ReVOS~\cite{yan2024visa} (Fig.~\ref{fig:suppl_qual_revos}), and ReasonVOS~\cite{bai2024videolisa} (Fig.~\ref{fig:suppl_qual_reasonvos}). These examples span a diverse range of expression types, from motion-based descriptions and spatial relationships in MeViS, to causal and comparative reasoning in ReVOS, to commonsense and hypothetical reasoning in ReasonVOS. Across all benchmarks, \ours consistently produces accurate mask tracks, demonstrating its effectiveness.

\label{sec:suppl_qual}
\input{fig/suppl_qual}

\section{Additional Implementation Details}
\label{sec:suppl_details}

% Implementation Details
\subsection{Ablation study configurations}
\label{sec:suppl_implementation}

All experiments are conducted via MLLM served through vLLM~\cite{kwon2023efficient}, 
with temperature set to 0.2 and a maximum of 8192 output tokens. 
Ablation experiments use Qwen3-VL-8B-Instruct~\cite{bai2025qwen3} as the MLLM backbone 
and are evaluated on the MeViS \textit{valid u} set.

% Used prompts
\subsection{Detailed prompts}
\label{sec:suppl_prompts}

\input{fig/suppl_prompts}

We provide the prompts used in the \ours pipeline in Figs.~\ref{fig:suppl_prompts_1},~\ref{fig:suppl_prompts_2} (Concept Extraction), Fig.~\ref{fig:suppl_prompts_appearance} (Appearance Tool), and Fig.~\ref{fig:suppl_prompts_pruning} (Iterative Spatio-temporal Pruning).

% Future Works
\section{Future Works}
\label{sec:suppl_future_works}
As stronger MLLMs such as Gemini-3-Pro~\cite{team2023gemini} continue to emerge, a promising direction is to substitute them into the pipeline, which requires no architectural modification due to the training-free and modular nature of \ours. To fully leverage their improved reasoning capabilities, exploring more sophisticated prompting strategies such as few-shot examples or structured chain-of-thought is another promising avenue for future work.

%%%%%%%%%%%%%%%%%%%%%%%%%%%%%%%%%%%%%%%%%%%%%%%%%%%%%%%%%%%%%%%%%%%%%%%%%%%%%%%%%%%%%%%%%%%%%%%%

% \clearpage

% \bibliography{main}
% \end{document}

%% file: tab/suppl_sam3_ablation.tex
\begin{table}[!h]
\centering
\renewcommand{\arraystretch}{0.96}

\caption{
    \textbf{Ablation on MLLM reasoning.} Replacing our MLLM-based reasoning pipeline with direct input of the referring expression into SAM3 leads to a substantial drop in segmentation  accuracy and a sharp increase in empty mask ratio, demonstrating that MLLM 
    reasoning is essential for handling the complexity of referring expressions.
\label{tab:suppl_sam3_ablation}
}
\vspace{-10pt}
\resizebox{0.9\textwidth}{!}{%
\begin{tabular}{L{2.5cm}|Z{4.0cm}|Z{2.0cm}Z{2.0cm}Z{2.2cm}}
\toprule
Methods & Empty Mask Ratio (\%) & \( \mathcal{J} \) & \( \mathcal{F} \) & \( \mathcal{J}\&\mathcal{F} \) \\
\midrule
SAM3~\cite{carion2025sam3} & 38.9 & 36.8 & 44.7 & 40.8 \\
\rowcolor{blue}
\textbf{\ours} & 3.8 (\textcolor{ForestGreen}{- 35.1}) & 67.1 (\textcolor{ForestGreen}{+ 30.3}) & 73.6 (\textcolor{ForestGreen}{+ 28.9}) & 70.3 (\textcolor{ForestGreen}{+ 29.5}) \\
\bottomrule
\end{tabular}%
}
\end{table}

%% file: fig/suppl_concept_extraction.tex
\begin{figure*}[!t]
    \centering
    \includegraphics[width=\textwidth]{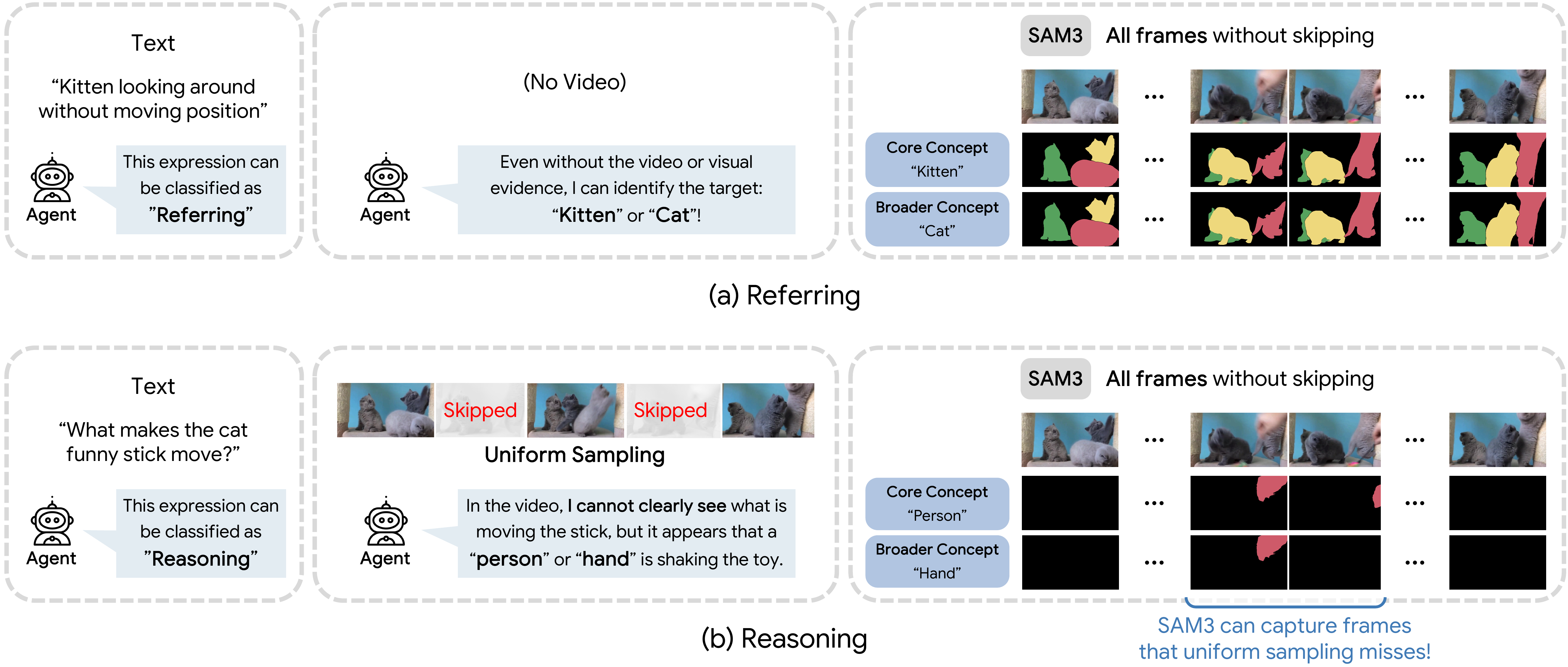}
    \caption{
        \textbf{Concept extraction for referring and reasoning queries.}
        For \textbf{referring} queries (a), the target is identifiable from
        the expression alone, so SAM3 is applied directly using the extracted
        concept.
        For \textbf{reasoning} queries (b), resolving the referent requires
        visual evidence, so the video is provided alongside the expression.
        Notably, SAM3 is applied across all frames rather than a sampled
        subset, enabling reliable detection even when the target object
        appears in only a small fraction of the video, as illustrated by
        the ``\textit{hand}'' visible in just two frames.
    }
    \label{fig:suppl_concept_extraction}
\end{figure*}

%% file: fig/suppl_appearance.tex
\begin{figure*}[t]
    \centering
    \includegraphics[width=\textwidth]{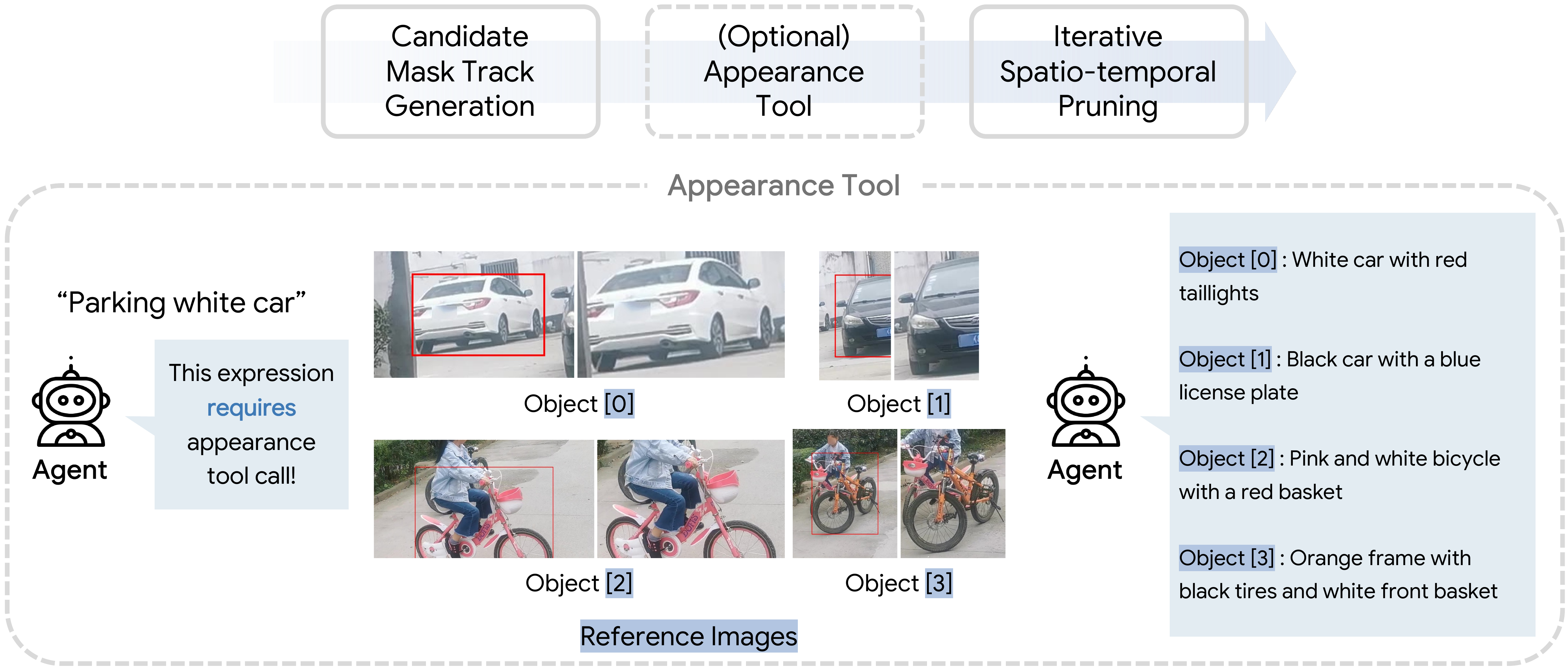}
    \caption{
        \textbf{Details of appearance tool.} Between candidate mask track generation and iterative spatio-temporal pruning, the appearance tool can be optionally invoked to obtain appearance information that may be obscured by visual prompting. When the agent determines that appearance-level evidence is needed for reasoning, it generates a brief phrase describing each candidate object's appearance.
    }
    \label{fig:suppl_appearance}
\end{figure*}

%% file: tab/suppl_appearance.tex
% ================= Table 1: Pipeline =================
\begin{table}[!t]
\centering
\renewcommand{\arraystretch}{0.96}

\caption{\textbf{Ablation for appearance tool call.}}
\label{tab:suppl_appearance}

\resizebox{0.5\textwidth}{!}{%
\begin{tabular}{L{4cm}|Z{1.0cm}Z{1.0cm}Z{1.2cm}}
\toprule
Component & \( \mathcal{J} \) & \( \mathcal{F} \) & \( \mathcal{J}\&\mathcal{F} \) \\
\midrule
w/o Appearance Tool & 64.6 & 70.9 & 67.7 \\
% \midrule
\rowcolor{blue}
\textbf{\ours} & 67.1 & 73.6 & 70.3 \\
\bottomrule
\end{tabular}%
}
\end{table}

%% file: tab/suppl_algorithm.tex
\begin{algorithm}[!h]
\caption{Candidate Mask Track Generation}
\label{alg_concept_extraction}
\begin{algorithmic}[1]
\Require Video $\mathcal{V} = \{I_t\}_{t=1}^{T}$, Language Query $Q$, MLLM $\mathcal{L}$, SAM3 $\mathcal{S}$
\Ensure Candidate Mask Tracks $M \in \{0,1\}^{T \times H \times W}$, Final selected concepts $C^\star$
\State $k \leftarrow 0$ \Comment{Iteration}
\State $M \leftarrow \emptyset$ \Comment{Mask tracks}
\State $Fail \leftarrow \emptyset$ \Comment{Failed Concept pairs}
\State $C^* \leftarrow \emptyset$ 

\While{$M = \emptyset \land k \leq k_{\max}$}
    \If{$k = 0$}
        \State $A^{(k)} \leftarrow \mathcal{L}(Q,\texttt{prompt}_{referring})$ \Comment{MLLM response}
        \If{$A^{(k)}[\text{query type}] = referring$}
            \State $C^{(k)} \leftarrow A^{(k)}[\text{concept pairs}]$
        \Else
            \State $A^{(k)} \leftarrow \mathcal{L}(\mathcal{V},Q,\texttt{prompt}_{reasoning})$ \Comment{Video conditioned extraction}
            \State $C^{(k)} \leftarrow A^{(k)}[\text{concept pairs}]$
        \EndIf
    \Else
        \State $A^{(k)} \leftarrow \mathcal{L}(\mathcal{V},Q,\texttt{prompt}_{reasoning}, Fail)$ \Comment{Retry with failed concepts}
        \State $C^{(k)} \leftarrow A^{(k)}[\text{concept pairs}]$
    \EndIf

    \For{each $(c_i^{core}, c_i^{broad}) \in C^{(k)}$}
        \State $M_i^{core} \leftarrow \mathcal{S}(\mathcal{V}, c_i^{core})$
        \State $M_i^{broad} \leftarrow \mathcal{S}(\mathcal{V}, c_i^{broad})$
        \State $c^{selected}_i \leftarrow \operatorname*{argmax}_{c \in \{c_i^{core}, c_i^{broad}\}} \textsc{Count}(M^c_i)$ \Comment{Concept selection}
        % \State $c^{selected}_i \leftarrow \operatorname*{argmax}\limits_{c \in \{c_i^{core}, c_i^{broad}\}} \textsc{Count}(c)$ \Comment{Select concept with more instances}
        \State $M_i^{selected} \leftarrow \mathcal{S}(\mathcal{V}, c_i^{selected})$ 
        \State $C^* \leftarrow C^* \cup c_i^{selected}$
        \State $M \leftarrow M \cup M_i^{selected}$ 
    \EndFor

    \If{$M = \emptyset$}
        \State $Fail \leftarrow Fail \cup C^{(k)}$
    \EndIf

    \State $k \leftarrow k + 1$
\EndWhile
\end{algorithmic}
\end{algorithm}

\begin{algorithm}[!h]
\caption{Appearance Tool}
\label{alg_appearance_tool}
\begin{algorithmic}[1]
\Require Language Query $Q$, MLLM $\mathcal{L}$, Reference images $I$
\Ensure Appearance information $\mathcal{A}$

\State $B \leftarrow \mathcal{L}(Q,\texttt{prompt}_{appearance\_requirement})$ \Comment{Check appearance requirement}
\If{$B$}
    \State $\mathcal{A} \leftarrow \mathcal{L}(I, Q, \texttt{prompt}_{appearance\_retrieval})$ \Comment{Extract appearance descriptions}
\Else
    \State $\mathcal{A} \leftarrow \emptyset$
\EndIf
\end{algorithmic}
\end{algorithm}

\begin{algorithm}[!h]
\caption{Iterative Spatio-Temporal Pruning}
\label{alg_selection}
\begin{algorithmic}[1]
\Require Video $\mathcal{V} = \{I_t\}_{t=1}^{T}$, Language Query $Q$, MLLM $\mathcal{L}$,  Candidate Mask Tracks $M \in \{0,1\}^{T \times H \times W}$, Appearance information $\mathcal{A}$, Selected Concepts $C^*$
\Ensure Predicted masks $\mathcal{M} \in \{0,1\}^{T \times H \times W}$

\State $M^{(0)} \leftarrow M$, \quad $\mathcal{T}^{(0)} \leftarrow \bigcup_{m_i \in M^{(0)}} \mathcal{T}(m_i)$, \quad  $\mathcal{M} \leftarrow \emptyset$
\For{$r = 0, 1, \ldots, r_{\max}$}
    \State Uniform sampled frames from $\mathcal{T}^{(r)}$
    \State $V^* \leftarrow visualize(\mathcal{V}, M^{(r)})$ \Comment{Visualize mask tracks}
    \State Classify each $m_i \in M^{(r)}$ as \texttt{Acc.}, \texttt{Rej.}, or \texttt{Unc.} via $\mathcal{L}(V^*, Q, \texttt{prompt}_{select},C^\star, \mathcal{A})$
    \State $\mathcal{M} \leftarrow \mathcal{M} \cup \{m_i \mid \texttt{Accepted}\}$
    \State $M^{(r+1)} \leftarrow \{m_i \in M^{(r)} \mid \texttt{Uncertain}\}$
    \State $\mathcal{T}^{(r+1)} \leftarrow \bigcup_{m_i \in M^{(r+1)}} \mathcal{T}(m_i)$ \Comment{Narrow temporal scope}
    \If{$M^{(r+1)} = \emptyset$}
        \textbf{break}
    \EndIf
\EndFor
\State \Return $\mathcal{M}$
\end{algorithmic}
\end{algorithm}

%% file: fig/suppl_overall_reasoning.tex
\begin{figure*}[!t]
    \centering
    \includegraphics[width=\textwidth]{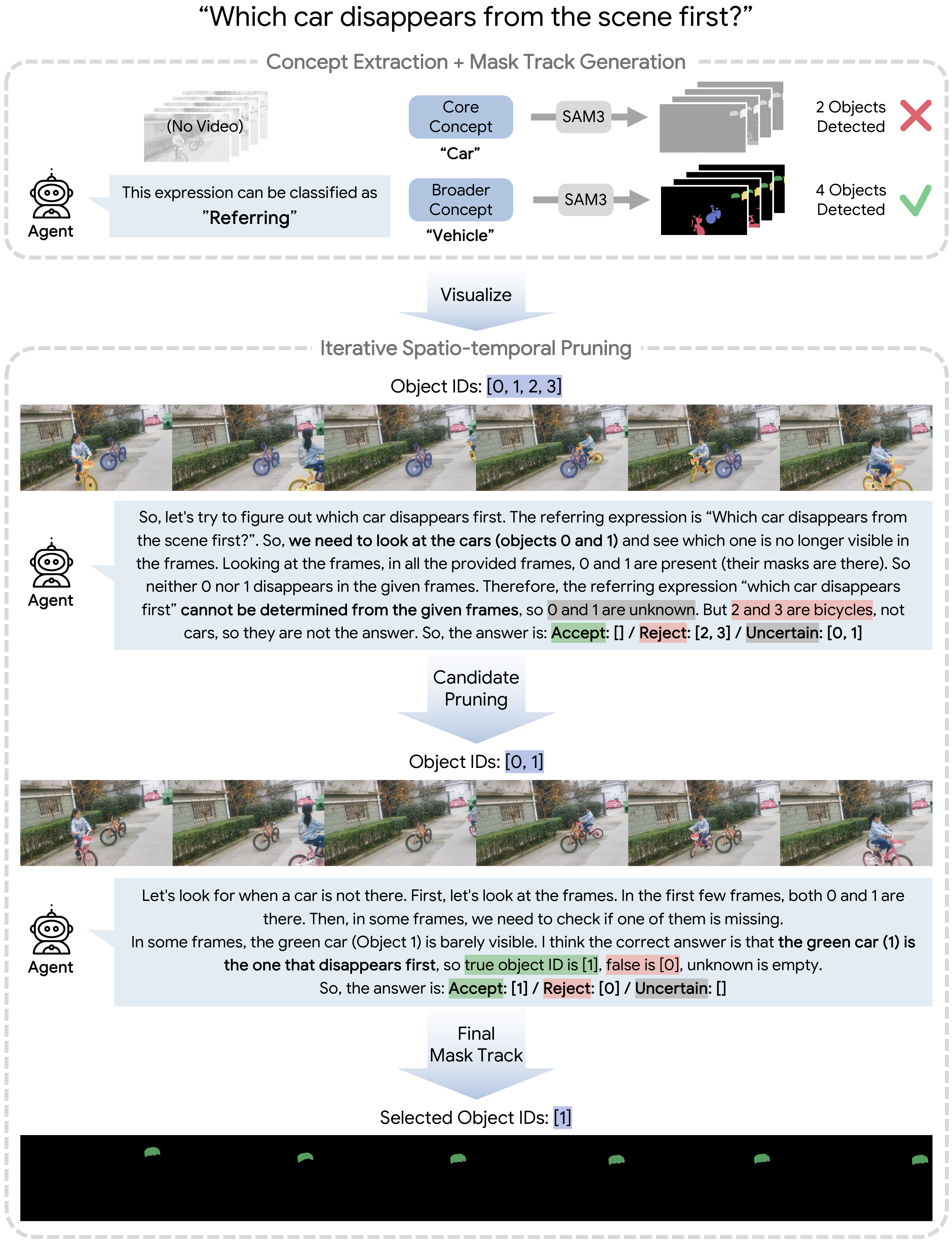}
    \caption{
    \textbf{Visualization of the overall pipeline of \ours.}
    This example illustrates a case where \textbf{candidate pruning} is applied.
    \ours progressively rejects irrelevant objects across iterations and identifies the correct target through proper iterative reasoning.
    }
    \label{fig:suppl_sample_1}
\end{figure*}

\begin{figure*}[!t]
    \centering
    \includegraphics[width=0.85\textwidth]{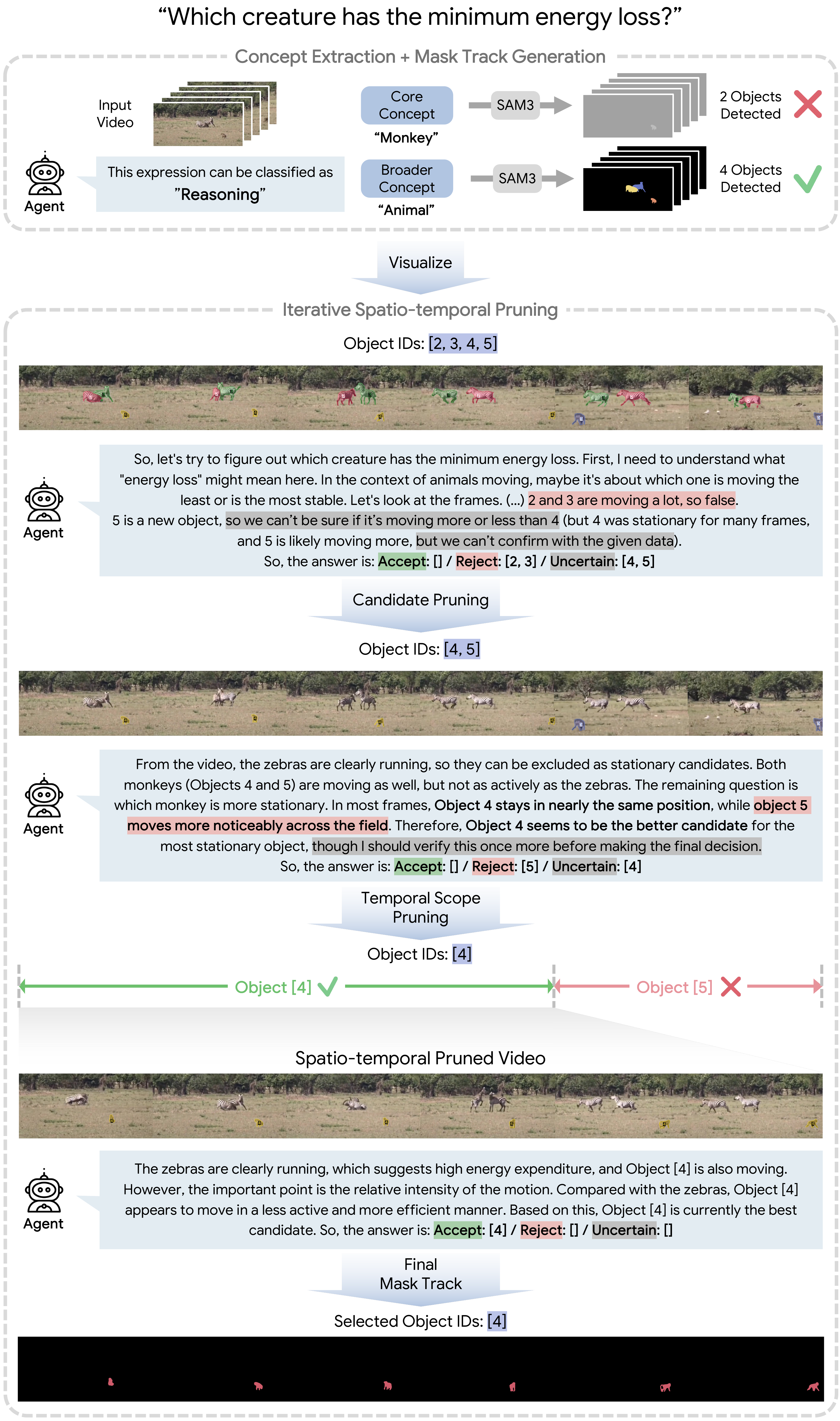}
    \caption{
    \textbf{Visualization of the overall pipeline of \ours.}
    This example illustrates a case requiring both \textbf{candidate pruning} and \textbf{temporal scope pruning}. After spatially narrowing the candidates, the pipeline further restricts the temporal scope to resolve the remaining ambiguity.
    }
    \label{fig:suppl_sample_2}
\end{figure*}

%% file: fig/suppl_qual.tex
\begin{figure*}[!t]
    \centering
    \includegraphics[width=\textwidth]{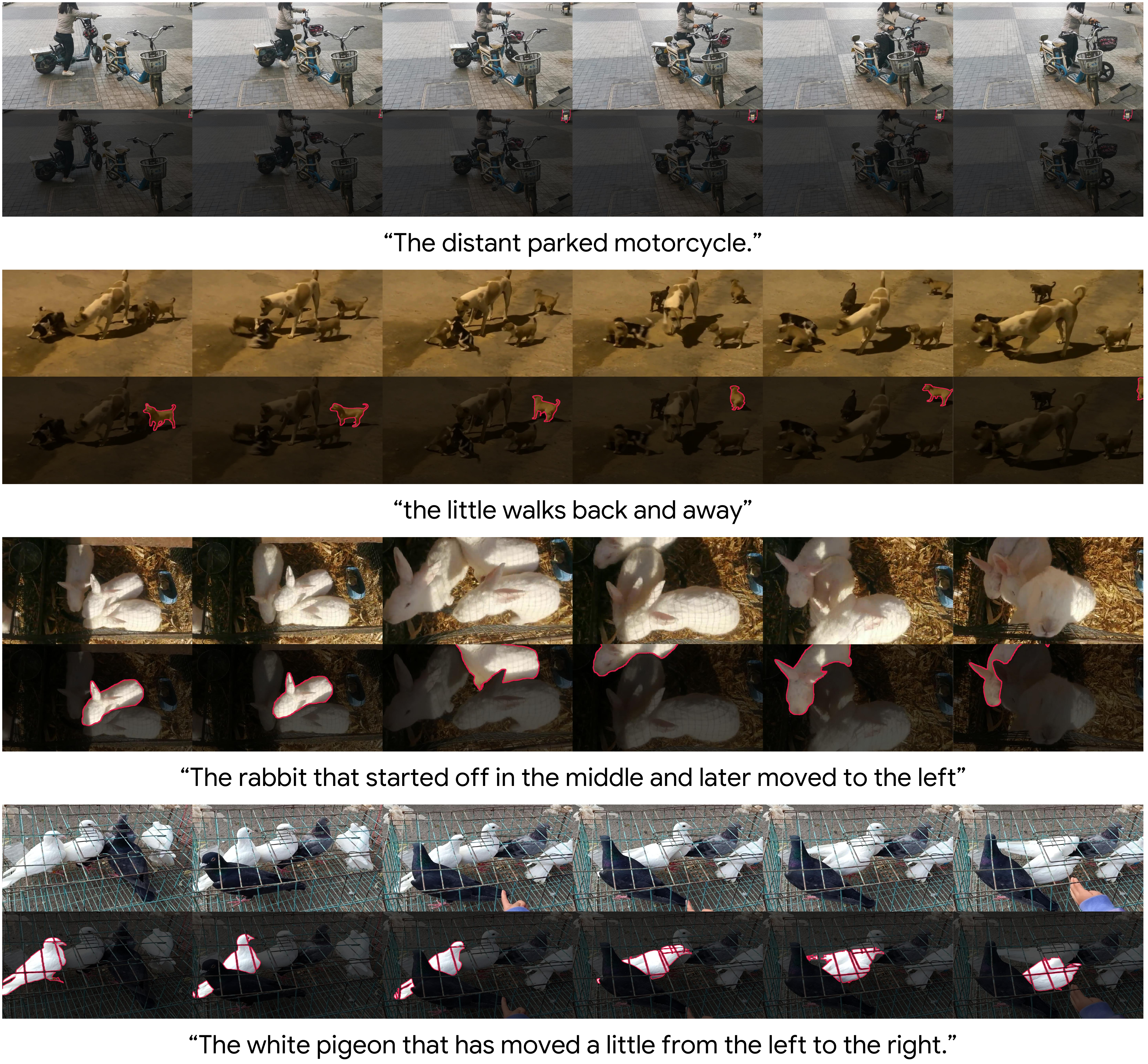}
    \caption{
        \textbf{Qualitative results in MeViS.}
        MeViS~\cite{ding2023mevis} expressions require understanding object motion and spatial relationships. \ours successfully identifies targets based on motion descriptions such as walking direction and relative displacement.
    }
    \label{fig:suppl_qual_mevis_1}
\end{figure*}

\begin{figure*}[!t]
    \centering
    \includegraphics[width=\textwidth]{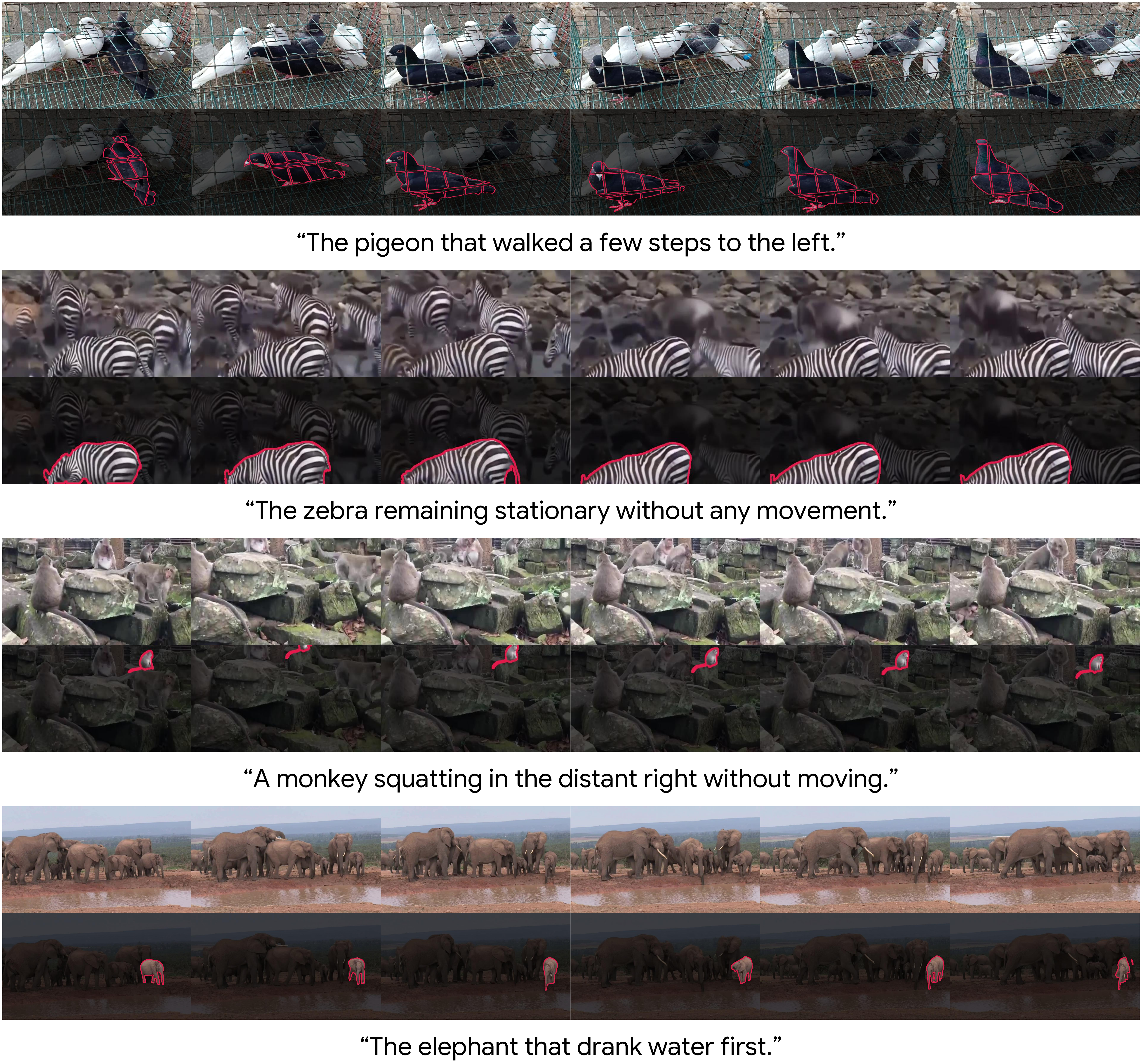}
    \caption{
        \textbf{Qualitative results in MeViS.}
        These examples involve distinguishing among multiple similar objects by their motion patterns, such as remaining stationary or performing a specific action first.
    }
    \label{fig:suppl_qual_mevis_2}
\end{figure*}

\begin{figure*}[!t]
    \centering
    \includegraphics[width=\textwidth]{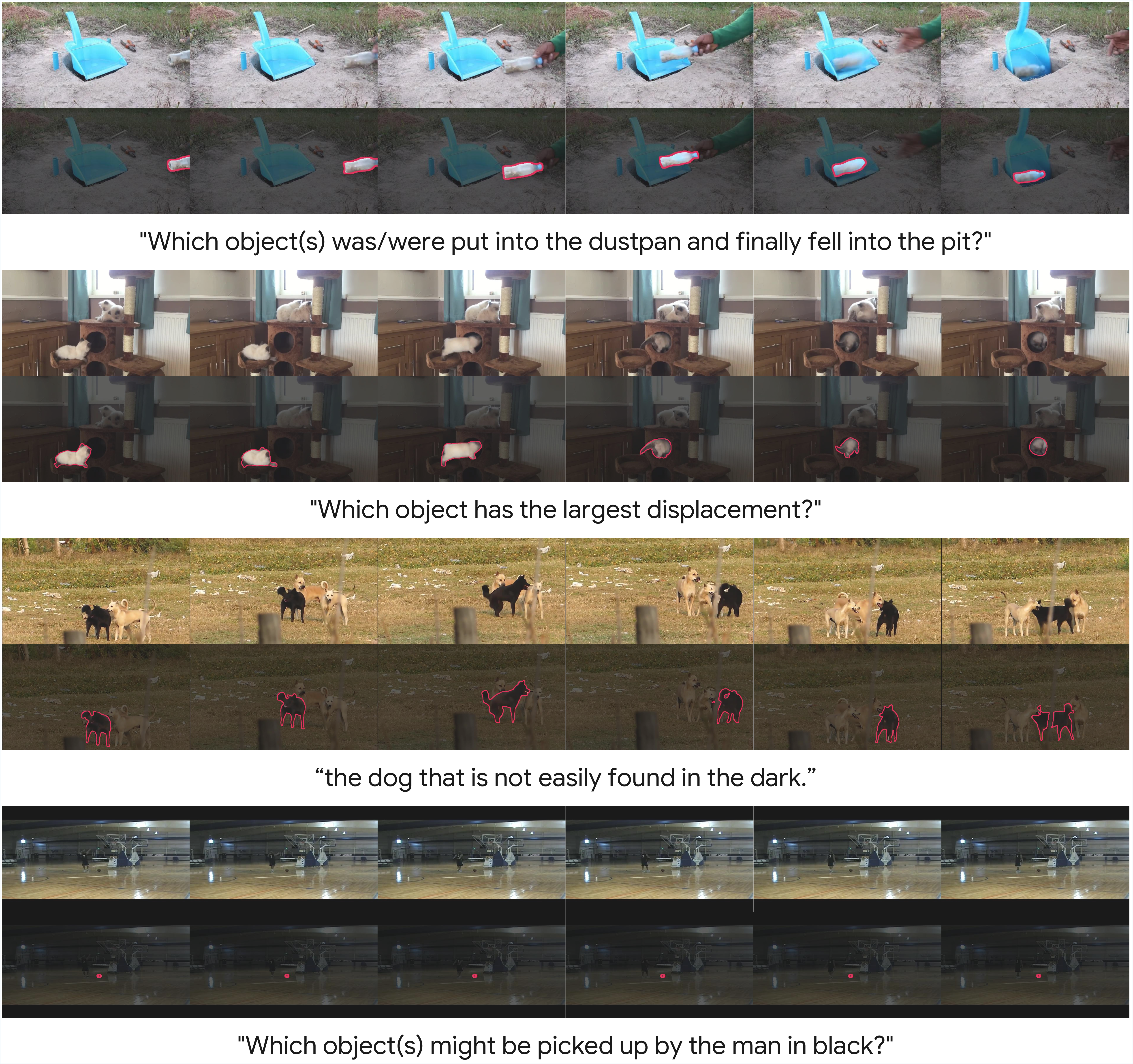}
    \caption{
        \textbf{Qualitative results in ReVOS.}
        ReVOS~\cite{yan2024visa} expressions often involve complex reasoning about object interactions and scene context. \ours correctly segments targets described through causal relationships, relative comparisons, and challenging visual conditions.
    }
    \label{fig:suppl_qual_revos}
\end{figure*}

\begin{figure*}[!t]
    \centering
    \includegraphics[width=\textwidth]{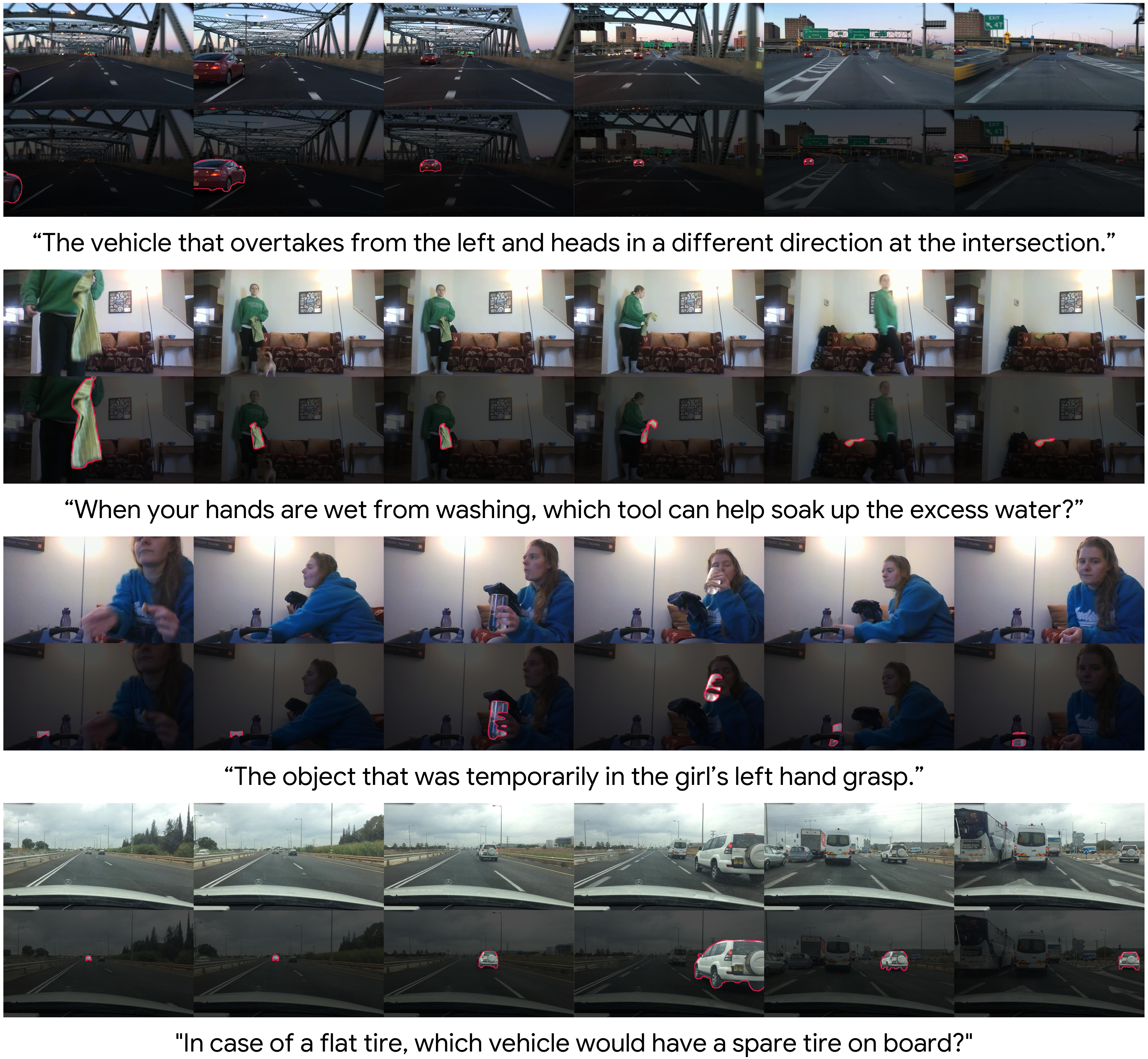}
    \caption{
        \textbf{Qualitative results in ReasonVOS.}
        ReasonVOS~\cite{bai2024videolisa} requires commonsense and world knowledge beyond direct visual cues. \ours handles expressions involving functional reasoning, hypothetical scenarios, and temporal event understanding.
    }
    \label{fig:suppl_qual_reasonvos}
\end{figure*}

%% file: fig/suppl_prompts.tex
\begin{figure*}[!t]
    \centering
    \includegraphics[width=0.8\textwidth]{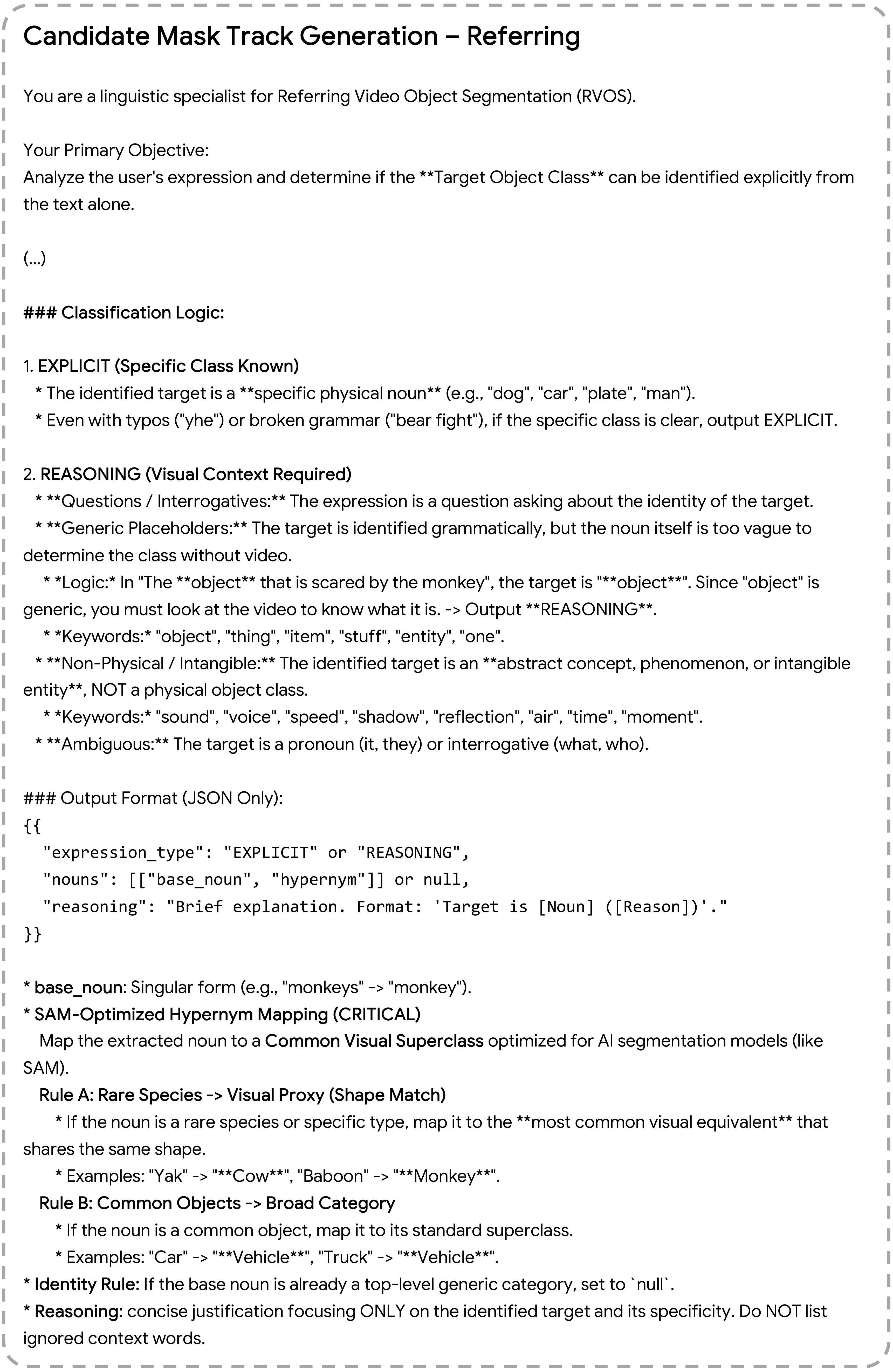}
    \caption{
        \textbf{Prompts used in \ours ($\texttt{prompt}_{referring}$).} The prompt for concept extraction,
        where the model receives a referring expression and extracts a core concept and a broader concept to guide SAM3 mask track generation.
    }
    \label{fig:suppl_prompts_1}
\end{figure*}

\begin{figure*}[!h]
    \centering
    \includegraphics[width=0.8\textwidth]{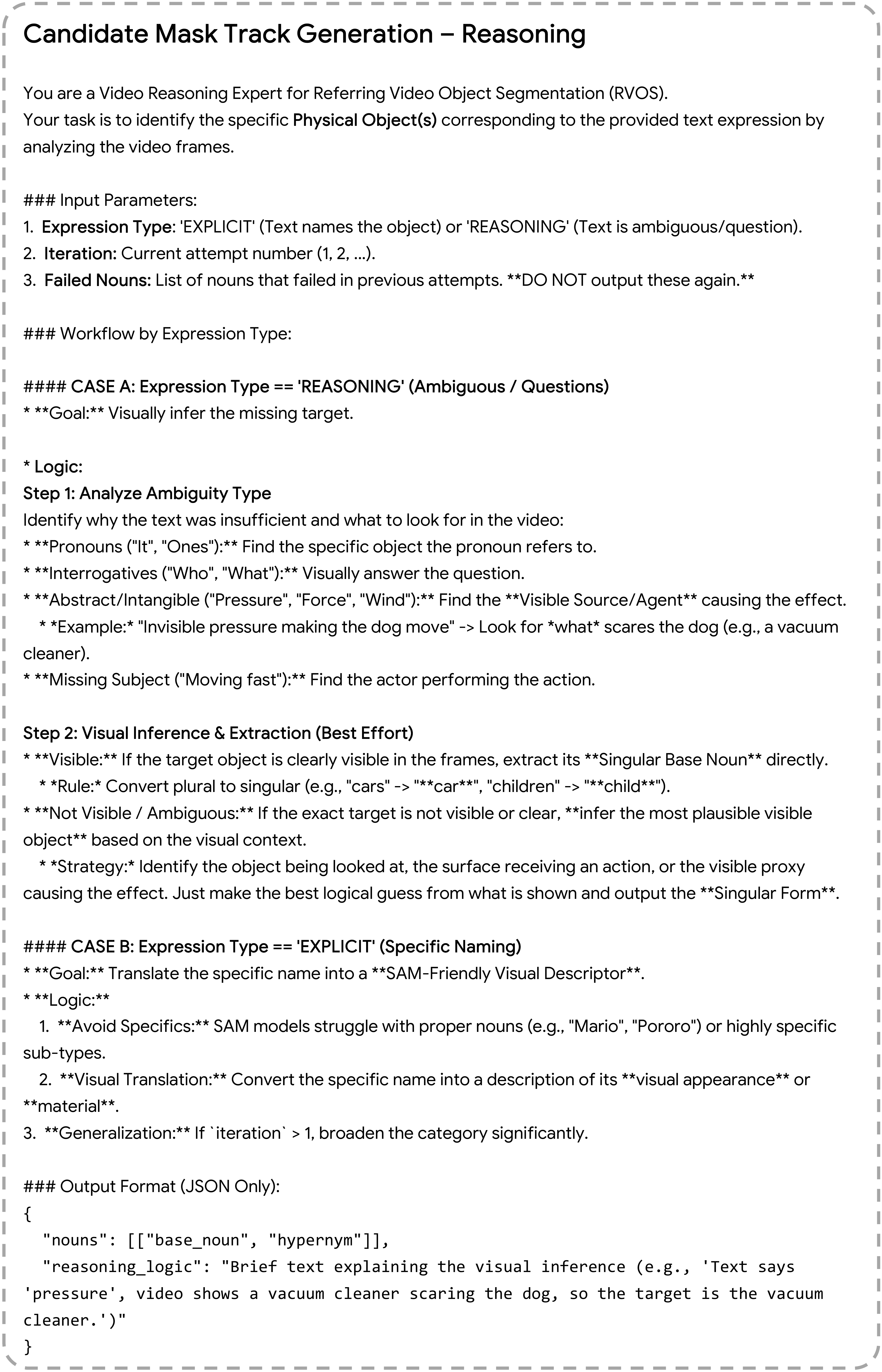}
    \caption{
        \textbf{Prompts used in \ours ($\texttt{prompt}_{reasoning}$).} The prompt for concept extraction, 
        where the model receives the referring expression and extracts a core concept and a broader concept to guide SAM3 mask track generation.
    }
    \label{fig:suppl_prompts_2}
\end{figure*}

\begin{figure*}[!h]
    \centering
    \includegraphics[width=0.8\textwidth]{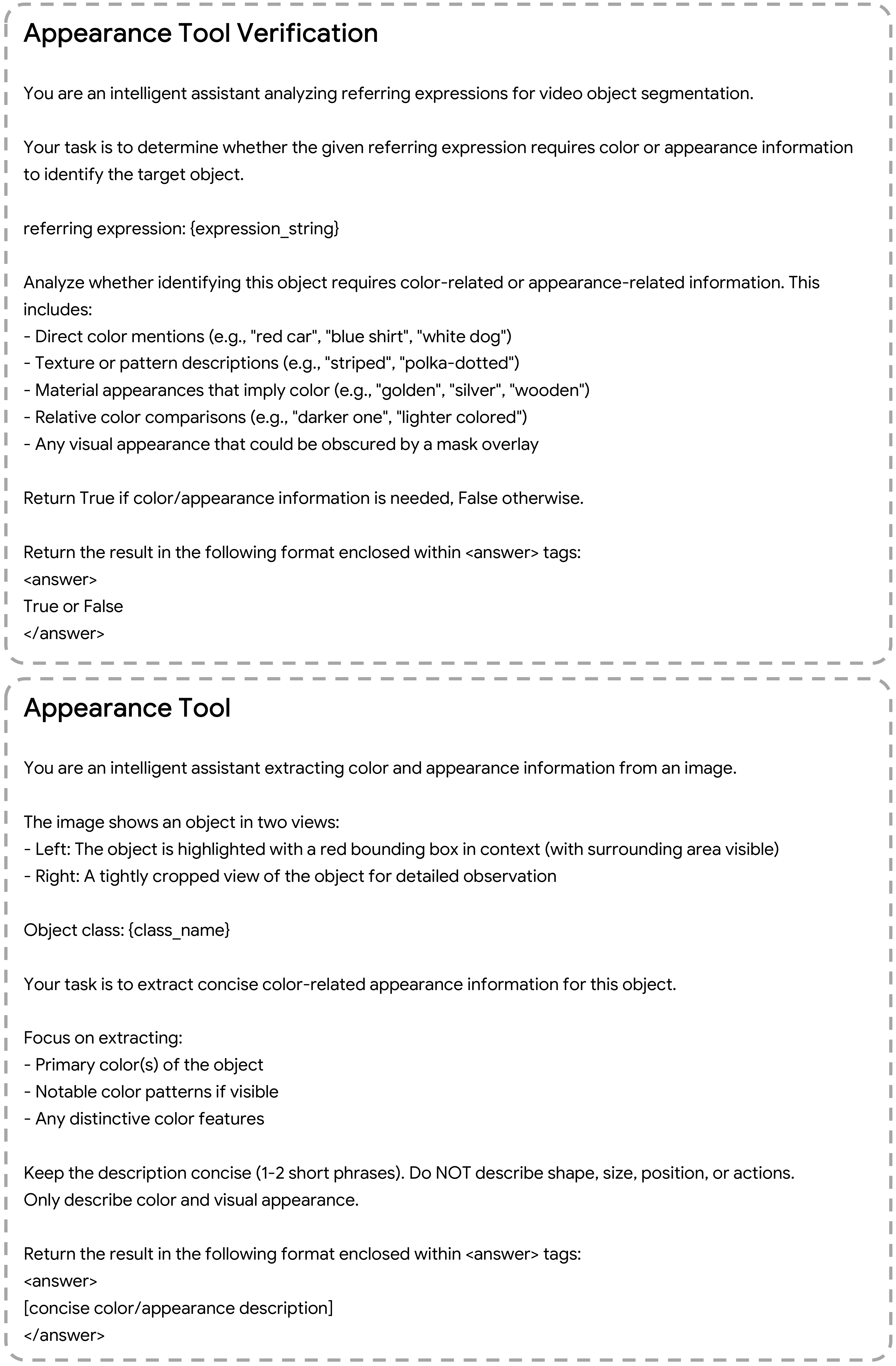}
    \caption{
    \textbf{Prompts used in \ours ($\texttt{prompt}_{appearance\_requirement}$} and \textbf{$\texttt{prompt}_{appearance\_retrieval}$).} The prompt for Appearance Tool Verification (top) determines whether the referring expression requires appearance-level evidence, and the prompt for Appearance Tool (bottom) extracts concise color and appearance descriptions from each candidate.
    }
    \label{fig:suppl_prompts_appearance}
\end{figure*}

\begin{figure*}[!h]
    \centering
    \includegraphics[width=\textwidth]{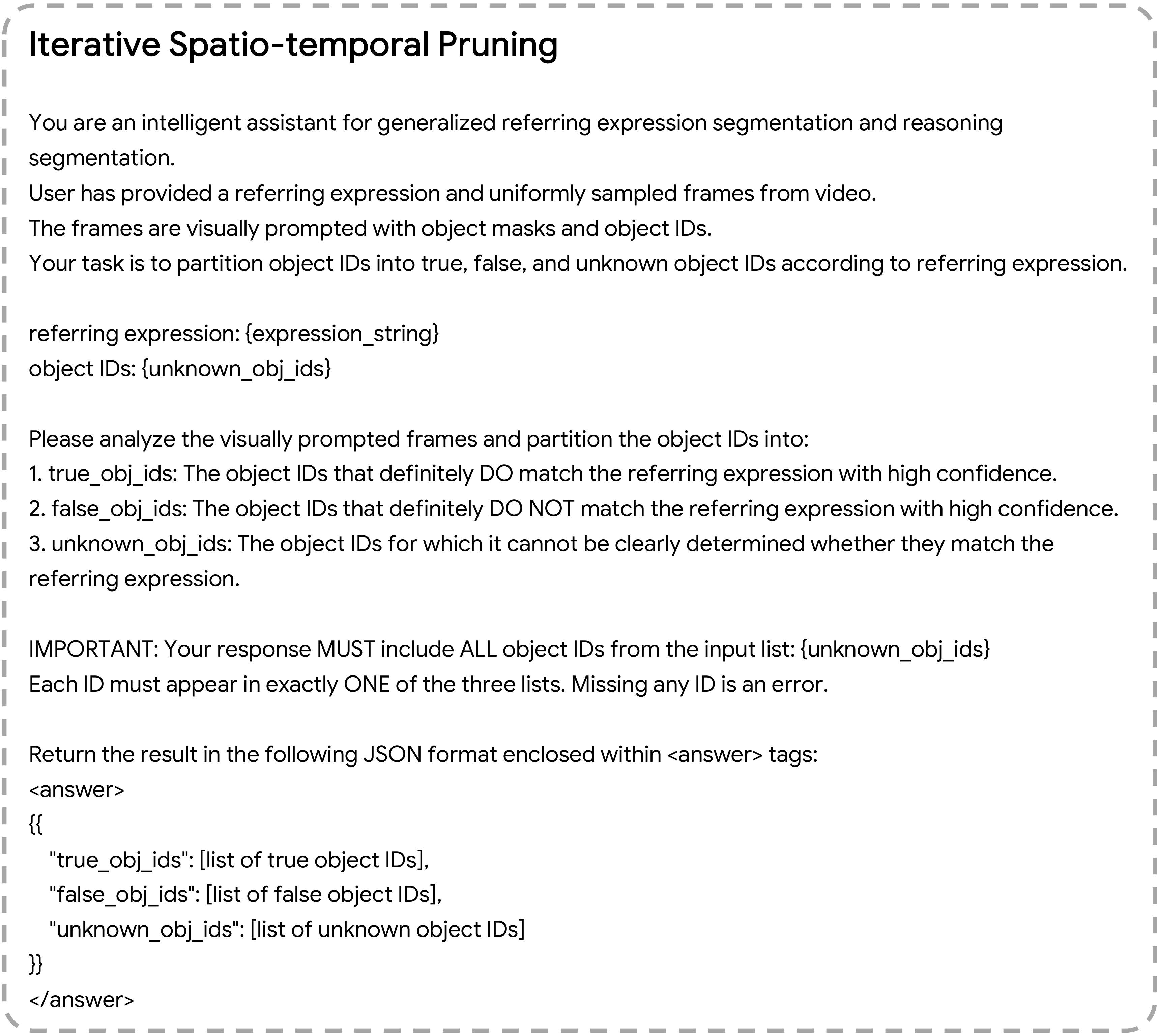}
    \caption{
        \textbf{Prompts used in \ours ($\texttt{prompt}_{select}$).} The prompt for Iterative Spatio-temporal Pruning,
        where the model receives a candidate mask track and a language query
        and classifies each candidate as \texttt{Accept}, \texttt{Reject}, or \texttt{Uncertain},
        progressively narrowing the candidate pool.
    }
    \label{fig:suppl_prompts_pruning}
\end{figure*}

%% file: main.bib
@String(ECCV  = {Eur. Conf. Comput. Vis.})

@String(AAAI  = {AAAI})

@String(ECCV  = {ECCV})

@inproceedings{gavrilyuk2018a2dsentences,
  title={Actor and action video segmentation from a sentence},
  author={Gavrilyuk, Kirill and Ghodrati, Amir and Li, Zhenyang and Snoek, Cees GM},
  booktitle={Proceedings of the IEEE conference on computer vision and pattern recognition},
  pages={5958--5966},
  year={2018}
}

@inproceedings{khoreva2018refvos,
  title={Video object segmentation with referring expressions},
  author={Khoreva, Anna and Rohrbach, Anna and Schiele, Brent},
  booktitle={Proceedings of the European Conference on Computer Vision (ECCV) Workshops},
  pages={0--0},
  year={2018}
}

@inproceedings{seo2020urvos,
  title={Urvos: Unified referring video object segmentation network with a large-scale benchmark},
  author={Seo, Seonguk and Lee, Joon-Young and Han, Bohyung},
  booktitle={European conference on computer vision},
  pages={208--223},
  year={2020},
  organization={Springer}
}

@inproceedings{wu2022referformer,
  title={Language as queries for referring video object segmentation},
  author={Wu, Jiannan and Jiang, Yi and Sun, Peize and Yuan, Zehuan and Luo, Ping},
  booktitle={Proceedings of the IEEE/CVF Conference on Computer Vision and Pattern Recognition},
  pages={4974--4984},
  year={2022}
}

@inproceedings{ding2023mevis,
  title={Mevis: A large-scale benchmark for video segmentation with motion expressions},
  author={Ding, Henghui and Liu, Chang and He, Shuting and Jiang, Xudong and Loy, Chen Change},
  booktitle={Proceedings of the IEEE/CVF international conference on computer vision},
  pages={2694--2703},
  year={2023}
}

@inproceedings{kirillov2023sam,
  title={Segment anything},
  author={Kirillov, Alexander and Mintun, Eric and Ravi, Nikhila and Mao, Hanzi and Rolland, Chloe and Gustafson, Laura and Xiao, Tete and Whitehead, Spencer and Berg, Alexander C and Lo, Wan-Yen and others},
  booktitle={Proceedings of the IEEE/CVF international conference on computer vision},
  pages={4015--4026},
  year={2023}
}

@inproceedings{ravi2024sam2,
  title={SAM 2: Segment Anything in Images and Videos},
  author={Ravi, Nikhila and Gabeur, Valentin and Hu, Yuan-Ting and Hu, Ronghang and Ryali, Chaitanya and Ma, Tengyu and Khedr, Haitham and R{\"a}dle, Roman and Rolland, Chloe and Gustafson, Laura and others},
  booktitle={The Thirteenth International Conference on Learning Representations},
  year={2024}
}

@article{carion2025sam3,
  title={Sam 3: Segment anything with concepts},
  author={Carion, Nicolas and Gustafson, Laura and Hu, Yuan-Ting and Debnath, Shoubhik and Hu, Ronghang and Suris, Didac and Ryali, Chaitanya and Alwala, Kalyan Vasudev and Khedr, Haitham and Huang, Andrew and others},
  journal={arXiv preprint arXiv:2511.16719},
  year={2025}
}

@inproceedings{huang2025alrefsam2,
  title={Unleashing the temporal-spatial reasoning capacity of gpt for training-free audio and language referenced video object segmentation},
  author={Huang, Shaofei and Ling, Rui and Li, Hongyu and Hui, Tianrui and Tang, Zongheng and Wei, Xiaoming and Han, Jizhong and Liu, Si},
  booktitle={Proceedings of the AAAI Conference on Artificial Intelligence},
  volume={39},
  number={4},
  pages={3715--3723},
  year={2025}
}

@inproceedings{lai2024lisa,
  title={Lisa: Reasoning segmentation via large language model},
  author={Lai, Xin and Tian, Zhuotao and Chen, Yukang and Li, Yanwei and Yuan, Yuhui and Liu, Shu and Jia, Jiaya},
  booktitle={Proceedings of the IEEE/CVF conference on computer vision and pattern recognition},
  pages={9579--9589},
  year={2024}
}

@inproceedings{yan2024visa,
  title={Visa: Reasoning video object segmentation via large language models},
  author={Yan, Cilin and Wang, Haochen and Yan, Shilin and Jiang, Xiaolong and Hu, Yao and Kang, Guoliang and Xie, Weidi and Gavves, Efstratios},
  booktitle={European Conference on Computer Vision},
  pages={98--115},
  year={2024},
  organization={Springer}
}

@article{bai2024videolisa,
  title={One token to seg them all: Language instructed reasoning segmentation in videos},
  author={Bai, Zechen and He, Tong and Mei, Haiyang and Wang, Pichao and Gao, Ziteng and Chen, Joya and Liu, Lei and Zhang, Zheng and Shou, Mike Z},
  journal={Advances in Neural Information Processing Systems},
  volume={37},
  pages={6833--6859},
  year={2024}
}

@article{yuan2025sa2va,
  title={Sa2va: Marrying sam2 with llava for dense grounded understanding of images and videos},
  author={Yuan, Haobo and Li, Xiangtai and Zhang, Tao and Sun, Yueyi and Huang, Zilong and Xu, Shilin and Ji, Shunping and Tong, Yunhai and Qi, Lu and Feng, Jiashi and others},
  journal={arXiv preprint arXiv:2501.04001},
  year={2025}
}

@inproceedings{xu2025videosegr1,
  title={Videoseg-r1: reasoning video object segmentation via reinforcement learning},
  author={Xu, Zishan and Guo, Yifu and Lu, Yuquan and Yang, Fengyu and Li, Junxin and Cai, Lihua},
  booktitle={Proceedings of the AAAI Conference on Artificial Intelligence},
  number={14},
  pages={11496--11504},
  year={2026}
}

@article{jin2025interrvos,
  title={InterRVOS: Interaction-aware Referring Video Object Segmentation},
  author={Jin, Woojeong and Kim, Seongchan and Lee, Jaeho and Kim, Seungryong},
  journal={arXiv preprint arXiv:2506.02356},
  year={2025}
}

@article{jiang2026referagent,
  title={Refer-Agent: A Collaborative Multi-Agent System with Reasoning and Reflection for Referring Video Object Segmentation},
  author={Jiang, Haichao and Liang, Tianming and Zheng, Wei-Shi and Hu, Jian-Fang},
  journal={arXiv preprint arXiv:2602.03595},
  year={2026}
}

@article{kao2025cotrvs,
  title={CoT-RVS: Zero-Shot Chain-of-Thought Reasoning Segmentation for Videos},
  author={Kao, Shiu-hong and Tai, Yu-Wing and Tang, Chi-Keung},
  journal={arXiv preprint arXiv:2505.18561},
  year={2025}
}

@article{liu2023visual,
  title={Visual instruction tuning},
  author={Liu, Haotian and Li, Chunyuan and Wu, Qingyang and Lee, Yong Jae},
  journal={Advances in neural information processing systems},
  volume={36},
  pages={34892--34916},
  year={2023}
}

@article{hurst2024gpt,
  title={Gpt-4o system card},
  author={Hurst, Aaron and Lerer, Adam and Goucher, Adam P and Perelman, Adam and Ramesh, Aditya and Clark, Aidan and Ostrow, AJ and Welihinda, Akila and Hayes, Alan and Radford, Alec and others},
  journal={arXiv preprint arXiv:2410.21276},
  year={2024}
}

@article{li2024llava,
  title={Llava-onevision: Easy visual task transfer},
  author={Li, Bo and Zhang, Yuanhan and Guo, Dong and Zhang, Renrui and Li, Feng and Zhang, Hao and Zhang, Kaichen and Zhang, Peiyuan and Li, Yanwei and Liu, Ziwei and others},
  journal={arXiv preprint arXiv:2408.03326},
  year={2024}
}

@article{team2023gemini,
  title={Gemini: a family of highly capable multimodal models},
  author={Team, Gemini and Anil, Rohan and Borgeaud, Sebastian and Alayrac, Jean-Baptiste and Yu, Jiahui and Soricut, Radu and Schalkwyk, Johan and Dai, Andrew M and Hauth, Anja and Millican, Katie and others},
  journal={arXiv preprint arXiv:2312.11805},
  year={2023}
}

@inproceedings{carion2020end,
  title={End-to-end object detection with transformers},
  author={Carion, Nicolas and Massa, Francisco and Synnaeve, Gabriel and Usunier, Nicolas and Kirillov, Alexander and Zagoruyko, Sergey},
  booktitle={European conference on computer vision},
  pages={213--229},
  year={2020},
  organization={Springer}
}

@inproceedings{radford2021learning,
  title={Learning transferable visual models from natural language supervision},
  author={Radford, Alec and Kim, Jong Wook and Hallacy, Chris and Ramesh, Aditya and Goh, Gabriel and Agarwal, Sandhini and Sastry, Girish and Askell, Amanda and Mishkin, Pamela and Clark, Jack and others},
  booktitle={International conference on machine learning},
  pages={8748--8763},
  year={2021},
  organization={PmLR}
}

@article{bai2025qwen3,
  title={Qwen3-vl technical report},
  author={Bai, Shuai and Cai, Yuxuan and Chen, Ruizhe and Chen, Keqin and Chen, Xionghui and Cheng, Zesen and Deng, Lianghao and Ding, Wei and Gao, Chang and Ge, Chunjiang and others},
  journal={arXiv preprint arXiv:2511.21631},
  year={2025}
}

@misc{he2024decouplingstatichierarchicalmotion,
      title={Decoupling Static and Hierarchical Motion Perception for Referring Video Segmentation}, 
      author={Shuting He and Henghui Ding},
      year={2024},
      eprint={2404.03645},
      archivePrefix={arXiv},
      primaryClass={cs.CV},
      url={https://arxiv.org/abs/2404.03645}, 
}

@article{miao2024htr,
  title={Temporally Consistent Referring Video Object Segmentation with Hybrid Memory},
  author={Miao, Bo and Bennamoun, Mohammed and Gao, Yongsheng and Shah, Mubarak and Mian, Ajmal},
  journal={IEEE Transactions on Circuits and Systems for Video Technology},
  year={2024},
  publisher={IEEE}
}

@article{chen2025expandingperformanceboundariesopensource,
  title={Expanding performance boundaries of open-source multimodal models with model, data, and test-time scaling},
  author={Chen, Zhe and Wang, Weiyun and Cao, Yue and Liu, Yangzhou and Gao, Zhangwei and Cui, Erfei and Zhu, Jinguo and Ye, Shenglong and Tian, Hao and Liu, Zhaoyang and others},
  journal={arXiv preprint arXiv:2412.05271},
  year={2024}
}

@misc{qwen2025qwen25technicalreport,
      title={Qwen2.5-VL Technical Report}, 
      author={Shuai Bai and Keqin Chen and Xuejing Liu and Jialin Wang and Wenbin Ge and Sibo Song and Kai Dang and Peng Wang and Shijie Wang and Jun Tang and Humen Zhong and Yuanzhi Zhu and Mingkun Yang and Zhaohai Li and Jianqiang Wan and Pengfei Wang and Wei Ding and Zheren Fu and Yiheng Xu and Jiabo Ye and Xi Zhang and Tianbao Xie and Zesen Cheng and Hang Zhang and Zhibo Yang and Haiyang Xu and Junyang Lin},
      year={2025},
      eprint={2502.13923},
      archivePrefix={arXiv},
      primaryClass={cs.CV},
      url={https://arxiv.org/abs/2502.13923},
}

@misc{li2025revsegincentivizingreasoningchain,
      title={ReVSeg: Incentivizing the Reasoning Chain for Video Segmentation with Reinforcement Learning}, 
      author={Yifan Li and Yingda Yin and Lingting Zhu and Weikai Chen and Shengju Qian and Xin Wang and Yanwei Fu},
      year={2025},
      eprint={2512.02835},
      archivePrefix={arXiv},
      primaryClass={cs.CV},
      url={https://arxiv.org/abs/2512.02835}, 
}

@article{yang2023set,
  title={Set-of-mark prompting unleashes extraordinary visual grounding in gpt-4v},
  author={Yang, Jianwei and Zhang, Hao and Li, Feng and Zou, Xueyan and Li, Chunyuan and Gao, Jianfeng},
  journal={arXiv preprint arXiv:2310.11441},
  year={2023}
}

@article{wei2024hyperseg,
  title={Hyperseg: Towards universal visual segmentation with large language model},
  author={Wei, Cong and Zhong, Yujie and Tan, Haoxian and Liu, Yong and Zhao, Zheng and Hu, Jie and Yang, Yujiu},
  journal={arXiv preprint arXiv:2411.17606},
  year={2024}
}

@inproceedings{wei2025instructseg,
  title={Instructseg: Unifying instructed visual segmentation with multi-modal large language models},
  author={Wei, Cong and Zhong, Yujie and Tan, Haoxian and Zeng, Yingsen and Liu, Yong and Wang, Hongfa and Yang, Yujiu},
  booktitle={Proceedings of the IEEE/CVF International Conference on Computer Vision},
  pages={20193--20203},
  year={2025}
}

@inproceedings{lin2025glus,
  title={Glus: Global-local reasoning unified into a single large language model for video segmentation},
  author={Lin, Lang and Yu, Xueyang and Pang, Ziqi and Wang, Yu-Xiong},
  booktitle={Proceedings of the Computer Vision and Pattern Recognition Conference},
  pages={8658--8667},
  year={2025}
}

@inproceedings{varma2023villa,
  title={Villa: Fine-grained vision-language representation learning from real-world data},
  author={Varma, Maya and Delbrouck, Jean-Benoit and Hooper, Sarah and Chaudhari, Akshay and Langlotz, Curtis},
  booktitle={Proceedings of the IEEE/CVF International Conference on Computer Vision},
  pages={22225--22235},
  year={2023}
}

@inproceedings{wang2025object,
  title={Object-centric video question answering with visual grounding and referring},
  author={Wang, Haochen and Chen, Qirui and Yan, Cilin and Cai, Jiayin and Jiang, Xiaolong and Hu, Yao and Xie, Weidi and Gavves, Stratis},
  booktitle={Proceedings of the IEEE/CVF International Conference on Computer Vision},
  pages={22274--22284},
  year={2025}
}

@inproceedings{gong2025devil,
  title={The devil is in temporal token: High quality video reasoning segmentation},
  author={Gong, Sitong and Zhuge, Yunzhi and Zhang, Lu and Yang, Zongxin and Zhang, Pingping and Lu, Huchuan},
  booktitle={Proceedings of the IEEE/CVF Conference on Computer Vision and Pattern Recognition},
  pages={29183--29192},
  year={2025}
}

@misc{openai2025gpt5,
  author       = {{OpenAI}},
  title        = {Introducing gpt-5},
  year         = {2025},
  month        = aug,
  note         = {August 7 2025}
}

@inproceedings{kwon2023efficient,
  title={Efficient memory management for large language model serving with pagedattention},
  author={Kwon, Woosuk and Li, Zhuohan and Zhuang, Siyuan and Sheng, Ying and Zheng, Lianmin and Yu, Cody Hao and Gonzalez, Joseph and Zhang, Hao and Stoica, Ion},
  booktitle={Proceedings of the 29th symposium on operating systems principles},
  pages={611--626},
  year={2023}
}

@misc{kao2026cotsegrethinkingsegmentationchainofthought,
      title={CoT-Seg: Rethinking Segmentation with Chain-of-Thought Reasoning and Self-Correction}, 
      author={Shiu-hong Kao and Chak Ho Huang and Huaiqian Liu and Yu-Wing Tai and Chi-Keung Tang},
      year={2026},
      eprint={2601.17420},
      archivePrefix={arXiv},
      primaryClass={cs.CV},
      url={https://arxiv.org/abs/2601.17420}, 
}

@misc{liu2025segzeroreasoningchainguidedsegmentation,
      title={Seg-Zero: Reasoning-Chain Guided Segmentation via Cognitive Reinforcement}, 
      author={Yuqi Liu and Bohao Peng and Zhisheng Zhong and Zihao Yue and Fanbin Lu and Bei Yu and Jiaya Jia},
      year={2025},
      eprint={2503.06520},
      archivePrefix={arXiv},
      primaryClass={cs.CV},
      url={https://arxiv.org/abs/2503.06520}, 
}
